\documentclass[a4paper,fleqn]{cas-dc}

\usepackage[numbers]{natbib}
\usepackage{amssymb}
\usepackage{enumitem}
\usepackage{amsmath}
\usepackage{xcolor}
\usepackage{float}
\usepackage{ulem}

\usepackage{verbatim}
\usepackage{multirow}
\usepackage{booktabs}
\usepackage{graphicx}
\usepackage{subfigure}
\usepackage{makecell}
\usepackage{placeins}

\begin{document}
\let\WriteBookmarks\relax
\def\floatpagepagefraction{1}
\def\textpagefraction{.001}

\title [mode = title]{{Fusing Transferred Priors and Physics-based Decomposition for Underwater Image Enhancement}}


\author[label1]{Haochen Hu}
\ead{haru-haochen.hu@connect.polyu.hk}
\author[label1]{Yanrui Bin}
\ead{binyanrui@gmail.com}
\author[label1]{Zhengyan Zhang}
\ead{sam-zhenyan.zhang@connect.polyu.hk}
\author[label1]{Minchen Wei}
\ead{minchen.wei@polyu.edu.hk}
\author[label1]{Chih-yung Wen}
\ead{cywen@polyu.edu.hk}
\author[label1]{Bing Wang\corref{cor1}}
\ead{bingwang@polyu.edu.hk}
\cortext[cor1]{Corresponding author}

\affiliation[label1]{organization={The Hong Kong Polytechnic University},
            city={Hung Hom, Kowloon},
            country={Hong Kong SAR}}


\begin{abstract}
The underwater images are captured within diverse water-medium conditions, leading to complex degradation, including color bias, low contrast, and blur effect. Recently, learning-based methods have demonstrated their potential for underwater image enhancement (UIE). However, most of the previous work focus on the training strategy or network design to make the enhanced result aligned well with the labels in datasets, ignoring that the labels are selected from the enhanced results of previous UIE methods and these pseudo-labels are noisy. Consequently, the performance of their models is not satisfactory to a certain extent. However, collecting the true labels of the underwater images is challenging. In this work, we propose a transfer learning-based UIE that does not require underwater images to have paired noisy or true labels for learning. Instead, the UIE task is first divided into global color correction, haze removal, and background noise suppression following the underwater physics. Then multiple types of prior from other vision tasks are leveraged as cross-domain supervision in each step. In this way, a novel UIE is available via transfer learning, and the physics-aligned UIE decomposition provides theoretical soundness. Qualitative and quantitative experiments demonstrate that our proposal based on physics and priors fusion achieves SOTA performance in the UIE task and effectively boosts downstream vision tasks, significantly outperforming benchmark methods. Project repo: \url{https://github.com/Haru2022/P2-UIE}.
\end{abstract}



\maketitle
\section{Introduction}

Underwater vision is essential for advancing marine biology research, enhancing underwater navigation systems, and improving the safety and efficiency of underwater construction and exploration activities. However, it is always degraded because of the water medium. Due to the apparent optical properties inherent to water bodies \cite{gonzalez2023survey}, light beams passing through the water suffer absorption and scattering, resulting in color bias, low contrast, and blur effects in the final underwater images. The degradation of underwater images complicates the execution of many visual algorithms \cite{cong2025pgf_if,li2025underwater_if,zheng2025novel_if}, which rely heavily on clear visual features or the correct color appearance, such as feature matching, edge detection, and saliency detection. Consequently, improving the quality of underwater images has attracted significant attention from computer vision communities. Numerous algorithms have been proposed over the past decade, achieving substantial progress. Among them, learning methods \cite{huang2023semiuir,li2019uieb, xie2024uveb,cong2023pugan,liu2024ccl,peng2025ssuie,pucci2025cevae,mei2025dpf,wang2025watercyclediffusion_if,hu2025pfusie} have demonstrated superiority over traditional handcrafted methods \cite{drews2016udcp,zhang2022mmle}.

\begin{figure*}[pos=!t]
  \centering
  \includegraphics[width=1.0\linewidth]{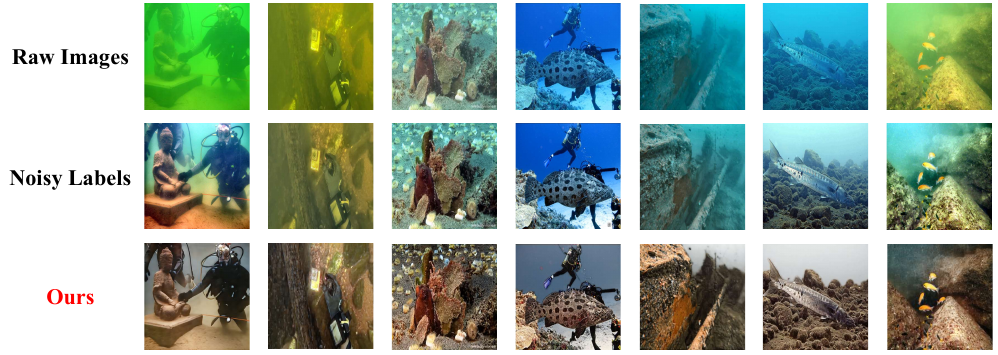}
  \caption{Example of the noisy labels in the previous paired dataset UIEB\cite{li2019uieb}. The first, second, and third row are the raw underwater images, labels in UIEB that still suffer low contrast and color bias, and the enhanced results of ours, respectively. }
  \label{labelbiascomp} 
\end{figure*}

Although much progress has been made with learning-based methods, the issue of performance degradation induced by biased datasets is still left to further research. Most of the recent learning-based methods are mainly based on supervised learning, which means the true label of degraded underwater images is needed. However, as it is not feasible to collect the ground truth of the underwater images by removing the water to capture images from the same perspective, the paired underwater datasets are always collected by manually selecting the labels from the previous work. For example, UIEB \cite{li2019uieb} and LSUI \cite{peng2023ushape} are two widely used paired datasets. To collect the label of raw underwater images, the authors first perform image enhancement using various previously proposed UIEs and then select the images with the best quality as corresponding labels. Quality evaluation is based on subjective scores of human volunteers. Although this strategy could integrate the abilities of different UIE methods to form a relatively reliable dataset, many labels still have weak quality, with almost unchanged degradation such as color bias and low contrast, as shown in Fig.\ref{labelbiascomp}. These pseudo-labels will mislead the UIE models and cause performance degradation. 

Nevertheless, there are few works that have focused on improving the quality of labels in underwater datasets in recent years, but pay too much attention to training strategy or network design to align well with labels. Some methods \cite{huang2023semiuir, li2020uwcnn, hu2025pfusie, zhou2024hclr} try to 1) introduce unsupervised contrast learning to use unpaired underwater images for training; 2) take image quality assessment (IQA) as a quantitative index to guide the model training; 3) use a fully synthetic dataset with ground truth for training, but all with limited performance improvement. Recently, related vision tasks have also seen advances in graph-based modeling\cite{li2025s2g_r1}, visual-textual mutual guidance\cite{liu2026visual_r1}, tensorized clustering\cite{wang2026tensorized_1_r2,wang2026tensorized_2_r2}, and cross-modality modulation\cite{peng2026bi_r2}, although these methods are not directly designed for underwater image enhancement.

In this work, we propose a new UIE method based on physics-aligned transfer learning. This solution does not require unbiased true labels of underwater images for training, but still achieves a significant performance improvement. The whole process is as follows: according to the Underwater Image Formation Model (UIFM), we first design a physics-based decomposition strategy, where the UIE is divided into global color correction, hazy removal, and background noise suppression. For global color correction, the diffusion prior embedded in the pre-trained diffusion model narrows the synthetic-to-real gap and consequently enables good performance on real-world images. In particular, the depth-related color bias is designed to degrade as two different constant values in the foreground and background regions. This setting, different from previous work, helps to further reduce the synthetic-real discrepancy. The underwater images processed by the first step will have roughly correct color pattern but show a hazy pattern. This is because the global-consistent color bias setting only helps to convert the channel-wise degradation into channel-irrelative degradation. According to UIFM, this residual of depth-related degradation is very similar to the atmosphere scattering model \cite{liu2021haze4k} that causes hazy images. Therefore, a paired real-world haze dataset \cite{ancuti2019densehaze} is utilized as the data prior to the removal of the residual effect. The real-world dense-hazy distribution would guide the model to eliminate the remaining hazy effect to get a realistic and clear foreground. Nevertheless, the background parts of processed images after hazy removal are always noisy because of the paradox between dense haze for high contrast and expected smooth background, which often makes the whole image visually unnatural. To further improve the quality of the enhancement, an additional step of background noise suppression is implemented based on the knowledge embedded in the semantic model \cite{ravi2024sam2,simeoni2025dinov3}, where the possible smooth and ideal background of images after global color correction is segmented and fused with the foreground part of the corresponding images after hazy removal. In this way, the final enhanced results will have a contrast-enhanced foreground part and a smooth (instead of noisy) background part, which meet the criteria of high-quality images. 

\FloatBarrier
\begin{figure}[pos=!t]
  \centering
  \includegraphics[width=1.0\linewidth]{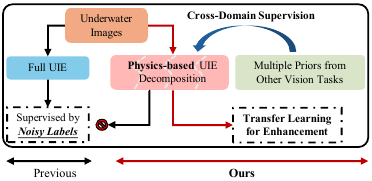}
  \caption{The Concept of our proposal. Instead of supervised learning by noisy labels, we resort to knowledge from relevant research fields for transfer learning via physics-based task decomposition.}
  \label{concept_illust} 
\end{figure}

To the best of our knowledge, this is the first time that multiple priors from relevant vision tasks are utilized for the cross-supervision of UIE. This transfer learning-based strategy as shown in Fig.\ref{concept_illust} provides a new perspective of the UIE solution that could avoid challenging label collection and the adversarial effect of noisy pseudo-label, but still can produce enhanced images that are much better than those of previous work. We think this is highly valuable. In addition, the proposed physical-aligned UIE decomposition strategy also ensures theoretical soundness, making each step and the corresponding fine-tuned model focus on the local enhancement to ease the learning difficulty. This general physics-based transfer learning cooperatively ensures the superior performance of the proposed P2UIE.

In summary, our main contributions are as follows.

\begin{itemize}
    \item An underwater image enhancement framework is designed that does not require paired labels of underwater image for learning, but still with high performance.
    \item For the first time, a transfer learning-based strategy is proposed to leverage multiple priors from relevant vision tasks for cross-domain supervision. This strategy avoids the adverse effect of heavily noisy pseudo-labels as well as the challenging collection of real-world accurate labels of the underwater images.
    \item A physics-based UIE decomposition strategy is also developed to provide theoretical soundness, so a satisfactory result is ensured. 
    \item We evaluated the proposed P2UIE on different datasets and performed a comprehensive analysis of the results quantitatively and qualitatively, demonstrating that P2UIE significantly outperforms other methods.
\end{itemize}

\section{Related Work}
\subsection{Underwater Image Formation Model}\label{uifm}

The optical properties of water bodies are determined by two factors: intrinsic optical properties (IOPs) and apparent optical properties (AOPs) \cite{gonzalez2023survey}. Together, IOPs and AOPs control how light behaves as it moves through water. For example, when a beam of light enters a given volume of water, it splits into three components: One portion is scattered in multiple directions, another is absorbed by the water, and the rest continues to travel through. Compared to the original beam, the transmitted light loses energy because of both scattering and absorption. The Underwater Image Formation Models (UIFMs) are all based on the above theory. McGlamery and Jaffe's model \cite{mcglamery1980computer, jaffe1990computer} proposed that in underwater environments, the incident light, whether from a light source or a certain point of an object surface, consists of two main components: direct light and forward-scattered light from nearby regions, assuming small-angle scattering. As a result, the total light energy $E_T$ captured by an imaging sensor is composed of three parts: $E_d$, the direct light coming from the object surface; $E_{fs}$ the object-related forward scattering; and $E_{bs}$, the background scattering light originating from the surrounding water:
\begin{equation}
E_T=E_d+E_{fs}+E_{bs}.
\end{equation}
Schechner and Karpel \cite{schechner2005recovery} extended the model to a widely referenced version of the UIFM definition as:
\begin{equation}
I_{total}=e^{f(\lambda,z)}(L_{object}(\lambda)+L_{object}(\lambda)\ast g_z)+B,
\end{equation}
where $\lambda$ denotes the spectrum, with $\lambda\in\{R,G,B\}$ if the spectrum is discretized. z is the object-to-sensor depth. $e^{f(\lambda,z)}$ represents the attenuation of light intensity due to absorption and scattering, where $f<0$. $L_{object}$ and $I_{total}$ are the unattenuated signal and the final underwater image captured by the sensor, respectively. $g_z$ is a point spread function representing the depth-related forward-scattering effect. The final term $B$ denotes background scattering (back-scattering), which is also depth-related:
\begin{equation}
    B=B_{\infty}(1-e^{f(\lambda,z)}).
\end{equation}
$B_{\infty}$ is the veiling light that is the accumulated back-scattering from the infinity background to the sensor. For a given water medium parameter, $B_{\infty}$ is constant. In \cite{chiang2011nerd}, the final captured underwater image $I^{uw}_{\lambda} (x)$ is formulated as:
\begin{equation}
I^{uw}_{\lambda} (x)=J_{\lambda}(x) \cdot t_{\lambda}(x) + B_{\lambda}(1-t_{\lambda}(x)), \label{nerd1}
\end{equation}
where $J_{\lambda}(x)$ is the scene radiance at a point $x$, $B_{\lambda}$ is the veiling light, and $\lambda$ means the three discrete color channels. $t_{\lambda}(x)$ is the residual energy ratio that denotes the degree of attenuation. It describes the ratio of the residual energy of a light beam reaching the camera and the initial energy of the light beam starting from the object:
\begin{equation}
    t_{\lambda}(x)=\frac{E_{\lambda}^{\mathrm{residual}}(x)}{E_{\lambda}^{\mathrm{initial}}(x)}=10^{-\beta(\lambda)d(x)}=\mathrm{Nrer}_\lambda^{d(x)}, \label{nerd2}
\end{equation}
where $d(x)$ is the object-to-camera depth, and $\mathrm{Nrer}_{\lambda}\in(0,1)$ (\textbf{N}ormalized \textbf{R}esidual \textbf{E}nergy \textbf{R}atio) is a constant under a given water condition representing the intrinsic properties of the water medium. Note that the scene radiance $J_{\lambda}(x)$ has already been attenuated by the water in the vertical direction from the surface of the water to the object. Thus, $J_{\lambda}(x)$ can be rewritten as:
\begin{equation}
    J_{\lambda}(x)=I_{\lambda}^{air}(x) \cdot \mathrm{Nrer}_\lambda^{D(x)}, \label{nerd3}
\end{equation}
where $I_{\lambda}^{air}(x)$ is the scene radiance captured in the air and $D(x)$ is the depth from the water surface to the object, respectively. Thus, according to the formulas \ref{nerd1}, \ref{nerd2}, \ref{nerd3}, the final UIFM is:
\begin{equation}
\begin{aligned}
    I^{uw}_{\lambda} (x)=&(I_{\lambda}^{air}(x) \cdot \mathrm{Nrer}_\lambda^{D(x)})\cdot \mathrm{Nrer}_\lambda^{d(x)}+ B_{\lambda}(1-\mathrm{Nrer}_\lambda^{d(x)})\\=&I_{\lambda}^{air}(x) \cdot  \mathrm{Nrer}_\lambda^{D(x)+d(x)} + B_{\lambda}(1-\mathrm{Nrer}_\lambda^{d(x)}). \label{nerd4}
\end{aligned}
\end{equation}

In summary, underwater images suffer from two main degradations: First, the channel-wise energy decay causes the color bias, which makes the images blue or green; second, a low contrast exists in many images because of the attenuated direct signal and the added scattering effect (dominated by the back-scattering).

\subsection{Underwater Image Enhancement Methods}

Image enhancement and restoration problems have long received significant attention from the community \cite{r12022aaai,r12024jas,r12025tip}as has the UIE problem. The UIE starts with hand-crafted methods. For example, white balance, histogram equalization, and the Gray-world assumption are some simple enhancement techniques. Recently, \cite{zhang2022mmle,zhuang2022laplacian} have been proposed based on retinex or color space theory. In MMLE \cite{zhang2022mmle}, images are decomposed into LAB color space where the reflectance component is sharpened while the illumination part is adjusted for contrast enhancement and color correction simultaneously, therefore achieving competitive results. However, hand-crafted methods require complex parameter design and struggle to extract knowledge from data. As a result, they have been gradually replaced by learning-based approaches.

Learning-based methods instead aim to solve the UIE problem in a data-driven manner. For physics-decoupling methods, water effects based on UIFM are estimated and decoupled to get target images, such as channel-wise attenuation coefficients, the veiling light, and the scene depth map \cite{hambarde2021uw,cong2023pugan,mei2025dpf}. Among them, PUGAN \cite{cong2023pugan} designs a Par-subnet to estimate physical parameters including depth map and attenuation coefficient. Together with the adversarial learning and degradation quantization module,  the enhanced result is acceptable. Similarly, DPF-Net \cite{mei2025dpf} also proposes a Degraded Parameters Estimation Module (DPEM) to guide the Physical Feature Generation Module (PFGM) for the following Feature Fusion. Non-physics decoupling methods are often developed to solve the problem in an end-to-end manner, focusing on new training strategies or network designs that can better extract the knowledge embedded in underwater image datasets where the raw underwater images and the corresponding enhanced images are paired for training \cite{huang2023semiuir,zhou2024hclr,pucci2025cevae,liu2024ccl,peng2025ssuie,hu2025pfusie}. HCLR \cite{zhou2024hclr} proposes to utilize hybrid contrastive learning regularization to move the enhanced image closer to the clear image, and CCL \cite{liu2024ccl} designs a cascade contrast learning strategy. WF-DIff \cite{zhao2024wfdiff} considers the UIE in the form of frequency analysis and processing. SS-UIE \cite{peng2025ssuie} and CE-VAE \cite{pucci2025cevae} are mainly involved in the new network development.

Although much progress has been achieved with learning-based methods, little attention has been paid to the critical issue that there are a considerable number of images with noisy labels in the real-world paired datasets. Take two widely used paired datasets, LSUI \cite{peng2023ushape} and UIEB \cite{li2019uieb}, as examples: in these two datasets, the authors first enhanced the raw underwater images using several previous UIE methods. The enhanced candidates for each raw underwater image are then scored by humans, and the result with the highest score is selected as the synthesized pseudo-label. As shown in Fig.\ref{labelbiascomp}, although this way of label collection equals integration of the ever-best performance of previous works, a large number of labels are noisy because all previous works are not able to process them well. Consequently, the models trained by these noisy labels inevitably suffer from irreducible error and have sub-optimal performance. In particular, unlike on-land label collections, it is very challenging to remove the water medium for label collection in underwater scenarios. An alternative is to use synthetic datasets for training \cite{li2020uwcnn, huang2023semiuir, hu2025pfusie} where the in-air images are true labels, while the corresponding underwater images are synthesized based on the UIFM. Nevertheless, the synthetic-to-real gap leads to certain performance degradation when models are trained with synthetic images but implemented on real-world underwater images \cite{hu2025pfusie}.

\subsection{Diffusion Models} \label{dm_relatedwork}

The Diffusion model\cite{rombach2022sd} is a generative model that is widely used for image tasks. Based on the diffusion probabilistic model proposed in \cite{ho2020ddpm}, it learns to generate target images by progressively denoising the latent space representation with added noise. It has been widely adopted because of its surprising efficiency and ability to generate detailed and diverse images in various visual tasks including super-resolution\cite{xiao2023diffappsr}, inpainting \cite{kim2025diffappinpainting}, and image restoration \cite{luo2025diffapprestoration}. More importantly, the open-source diffusion models are always pre-trained with large-scale datasets across various domains. It means that the pre-trained models already have extensive real-scene understanding, which effectively reduces the time and effort required to extend the diffusion models to new applications. Recently, PFUSIE \cite{hu2025pfusie} proposes to leverage embedded real-world knowledge in the pre-trained diffusion model, which is also referred to diffusion prior, to narrow the synthetic-to-real gap so that diffusion-based UIE models trained by the fully synthetic dataset show competitive performance on the real-world UIE task. However, it is still difficult for the diffusion model to work perfectly with real and complex underwater physics by using fully synthetic datasets based on simplified UIFM.

\section{Methodology}\label{sec_method}

\subsection{Preliminary} \label{preliminary section}

\begin{figure}[pos=!t]
  \centering
  \includegraphics[width=1.0\linewidth]{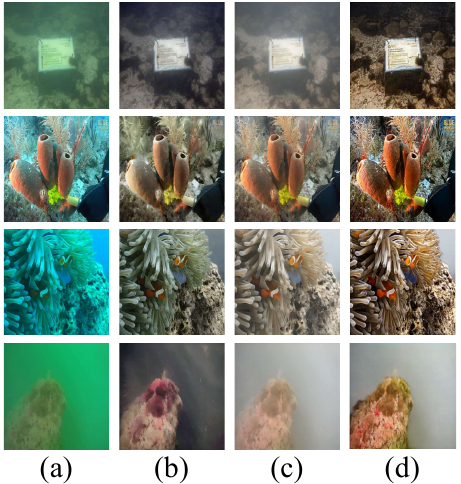}
  \caption{The limitation of finetuning diffusion model by fully synthetic underwater images. (a), (b), (c), (d) are the raw underwater images, enhanced results by \cite{hu2025pfusie}, the intermediate enhanced result by the global color correction, and the final enhanced result of our overall proposal, respectively.}
  \label{diffsyn_drawback} 
\end{figure}

Although diffusion prior helps much to narrow the synthetic-to-real gap, the diffusion model learned directly from full synthetic underwater images based on the UIFM still suffers from certain performance degradation when implemented on the real-world UIE task, causing artifacts in some results. This is because the UIFM is a simplified version of the complex real underwater physics. As shown in Fig.\ref{diffsyn_drawback}(b), PFUSIE \cite{hu2025pfusie} struggles to obtain clear images that preserve the richness and correctness of the color pattern.

Therefore, instead of training a UIE model supervised by real world images with noisy labels or by synthetic images based on simplified UIFM, we propose to decompose the full physical effect into different sub-degradations and then take advantage of multiple model and data priors from related visual tasks that can provide labels or guidance for enhancement which are more reliable by virtue of being better-aligned with the real world. Taking formula \ref{nerd4} as an example, the effect could be roughly decomposed into 1) color-related degradation caused by the parameter $\mathrm{Nrer}_\lambda$ and 2) left depth-related degradation caused by $D(x)$ and $d(x)$. By such a physics-based model decomposition, different enhanced models focus only on the local simplified water effect, and therefore perform well on the local enhancement with the help of corresponding priors. In this way of physics-based transfer learning, the synthetic-real discrepancy resulting from simplified UIFM is significantly reduced.

\begin{figure*}[pos=!t]
  \centering
  \includegraphics[width=0.9\linewidth]{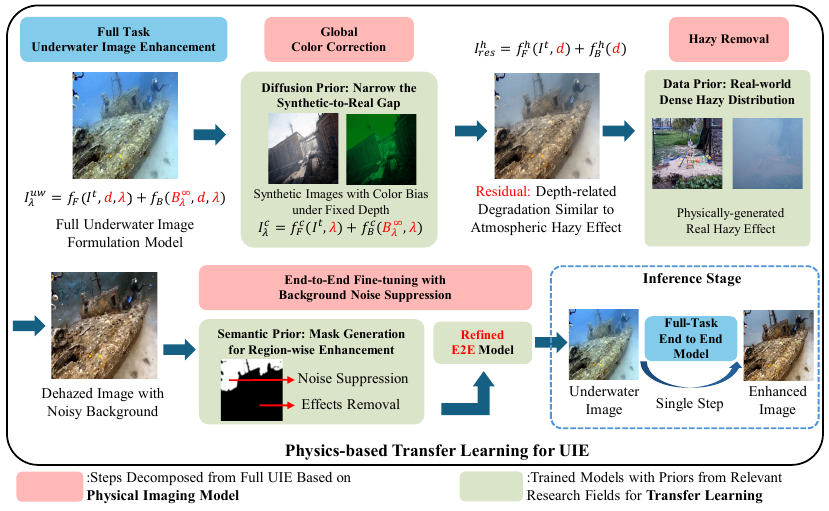}
  \caption{Framework of the Proposal.}
  \label{framework} 
\end{figure*}

Fig.\ref{framework} illustrates the general structure of our proposal about how different priors are integrated into each decomposition step. Generally speaking, these priors are primarily used to provide domain-specific adaption (pre-trained diffusion model) or supervisory signals (real hazy dataset and semantic model) to the targeting model. As mentioned above, the UIE task is mainly decomposed into three cascade steps based on the physical image formulation model, which are global color correction, haze removal, and final background noise suppression. First, for global color correction, synthetic images with region-wise constant color bias are used for training. This simple bias pattern along with the diffusion prior \cite{hu2025pfusie} allows the expected rough color correction in the real-world underwater images even trained in the synthetic domain. Next, for haze removal, another diffusion model is trained using a real-world dense haze dataset \cite{ancuti2019densehaze} where the haze effect is physically generated. The real distribution of the dense haze contributes to a clear and natural foreground result. After that, to suppress the possible noisy background part, the smooth background part of the result after global color correction is segmented semantic models SAM2 \cite{ravi2024sam2} and DINOv3 \cite{simeoni2025dinov3} and used for region-wise enhancement incorporated with the contrast-enhanced foreground part after hazy removal. The pre-trained diffusion model \cite{rombach2022sd} is used for color correction, hazy removal, and end-to-end training.

In the inference stage, only the refined model of the final end-to-end training is used for single-step underwater image enhancement.

\subsection{Practical Implementation}
\subsubsection{Parameter Configuration and Priors Implementation}

The water body is assumed to be homogeneous, implying constant absorption and scattering coefficients per unit voxel. This common assumption, widely adopted in physics-based underwater enhancement methods, is applicable to the vast majority of practical scenarios. The physics parameters for synthesis follow Equation \ref{nerd4}. The surface-to-object depth and object-to-sensor distance are randomly sampled to generate images with a continuous spectrum of visual clarity, ranging from clear to heavily blurred. In addition, the parameter range is expanded from several discrete values to a continuous range \cite{hu2025pfusie} to ensure that the proposal can be adopted under various color biases. Therefore, the generalizability of the model is guaranteed.

In summary, the priors used for transfer learning serve as cross-domain supervision signals or model parameter initialization: 1) The diffusion prior provides the initialized model parameters that are learned from Internet-scale datasets, and therefore provides extensive real scene understanding as analyzed in \cite{hu2025pfusie}. This will reduce training time and effort to fine-tune them to accommodate other tasks related to scene understanding. In particular, it helps the color correction step because there is no real-world dataset with true labels, and the model is trained by synthetic underwater images. Although the diffusion model is fine-tuned by synthetic data, it performs well in the task of real-world image color correction due to the embedded diffusion prior. 2) The real-world hazy images with physically-generated true labels and the foreground-background segmentation labels derived from pre-trained segmentation model provide the guidance for the supervision transfer learning, respectively. These labels are more aligned with real-world scenarios compared with those of full-synthesized based on simplified UIFM. 3) Similarly, the semantic models provide foreground-background labels to supervise the final end-to-end training aimed at smoothing the noisy background. Consequently, when these priors are integrated, the final model becomes more capable of handling the enhancement of the overall water effect. 

\subsubsection{Global Color Correction with Region-wise Color Bias}\label{sec_color_correction}

\begin{figure*}[pos=!t]
  \centering
  \includegraphics[width=0.7\linewidth]{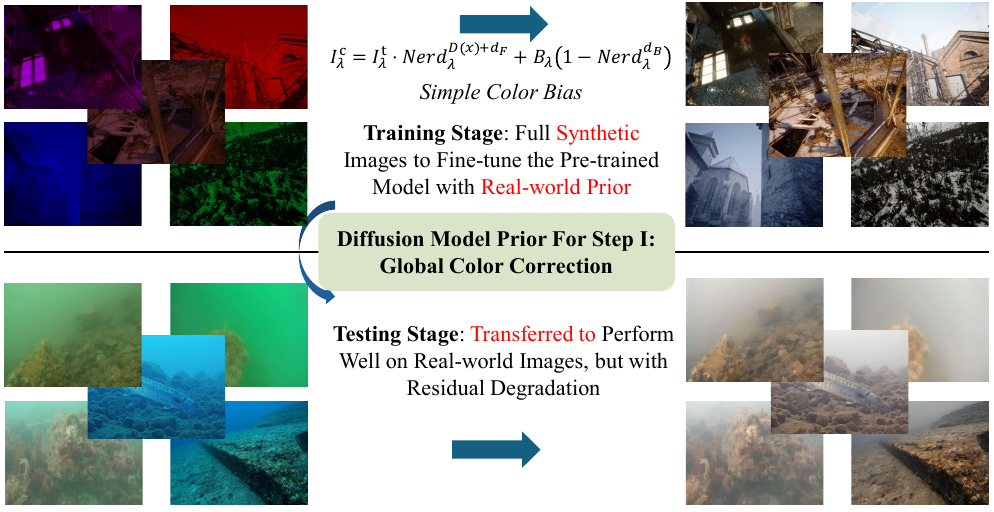}
  \caption{Step 1: Global Color Correction.}
  \label{task1} 
\end{figure*}

\begin{table}[pos=!t]
\centering
\begin{tabular}{|c|p{6cm}|}
\hline
\multicolumn{1}{|c|}{\textbf{Variables}} &
\multicolumn{1}{c|}{\textbf{Definitions}} \\
\hline
\hline
$\lambda$  & color channel in $R,G,B$ \\
$\mathrm{Nrer}$  & normalized residual energy ratio representing the energy decay in the water medium\\
$D(x)$ & surface-to-object (vertical) depth \\
$B_\lambda$ & channel-wise veiling light representing the accumulated back-scattering effect\\
$d_F, d_B$ & object-to-sensor (horizontal) depth of the foreground/background part \\
$I^h_{res}$ & residual degraded image after global color correction which showing haze effect \\
$f_F, f_B$ & residual degradation function for the foreground/background part after global color correction \\
$I^{clear}$ & clear image without any hazy effect \\
$I^{haze}$ & hazy image based on atmosphere scattering model \\
$T$ & transmission map representing the depth-depended degradation \\
$A$ & global atmospheric light \\
$F^h, F^c$ & semantic feature of the haze result after global color correction/the coarse result after hazy removal \\
$\mathcal{E}$ & encoder of the Variational AutoEncoder (VAE) to compress the input into latent space\\
$\mathcal{D}$ & decoder of the VAE to decompress the latent result into the pixel space\\
\hline
\end{tabular}
\caption{Important variables and definitions used in Section.\ref{sec_method}.}
\label{variable_def}
\end{table}

For ease of understanding, the definitions of important variables used in Section.\ref{sec_method} are given in Tab.\ref{variable_def}. In this step, a pre-trained stable diffusion model \cite{rombach2022sd} is fine-tuned to roughly correct the color bias of the raw underwater images, which is demonstrated in Fig.\ref{task1}. As analyzed in Section \ref{preliminary section}, the foreground part is set to have a fixed degradation depth $d_1$, while the background part is set to have another fixed degradation depth $d_2>d_1$. In this way, the learning burden is significantly reduced with less synthetic-real discrepancy. As shown in Fig.\ref{diffsyn_drawback}, PFUSIE \cite{hu2025pfusie} struggles to obtain clear images that preserve the richness and correctness of the color pattern (Fig.\ref{diffsyn_drawback}(b)). Instead, the proposed color correction only focuses on and thus performs well to roughly eliminate global color bias (Fig.\ref{diffsyn_drawback}(c)), leaving the left effect to the following steps for a final enhanced result with better quality (Fig.\ref{diffsyn_drawback}(d)). The simplified synthetic formula is described as follows:
\begin{equation}
I_{\lambda}^{c}=I_{\lambda}^{t}\cdot \mathrm{Nrer}_{\lambda}^{{D(x)+d_F}}+B_\lambda(1-\mathrm{Nrer}_{\lambda}^{d_B}), \label{nerd_color}
\end{equation}
where $I_{\lambda}^{c},I_{\lambda}^{t}$ are the synthetic images with region-wise color bias and corresponding target ground truth, $\mathrm{Nrer}$ is the water type parameter representing the level of degradation. $D(x)$ is the surface-to-object depth to simulate the global vertical color degradation. $d_F=d_1$ is the fixed depth for the foreground part. $d_B=d_1$ if the pixel belongs to the foreground part, while $d_B=d_2$ if the pixel belongs to the background part. $\lambda\in\{R,G,B\}$ means that depth-related degradation differs from color channels.

After global color correction, the enhanced images will show a haze effect. This is because removing the color bias defined by region-wise fixed depth serves to change the channel-wise degradation to be channel insensitive to $\lambda\in\{R,G,B\}$, but the effect caused by the scene-related depth distribution still exists. By comparing formula \ref{nerd4} and formula \ref{nerd_color}, which are the full UIFM and image formation model for global color bias, the residual effects could be roughly described as:
\begin{equation}
I_{res}^h(x)=I^{t}(x)\cdot f_F(x,d(x))+(1-f_B(x,d(x))),
\end{equation}
where $I_{res}^h$ is the residual degraded image, $f_F,f_B$ are the left degradation for the foreground part and the background part, which are depth-related but channel-insensitive, and $x$ denotes pixels. $d(x)$ means that the depth is related to pixels. This type of degradation is essentially very similar to the atmospheric dehazing equation \cite{liu2021haze4k} with white-like scattering medium causing hazy effect:
\begin{equation}
    I^{haze}=I^{clear}\cdot T(x,d(x))+A(1-T(x,d(x))),
\end{equation}
where T is the transmission map that represents the depth-dependent factor and affects the fraction of light preserved, A$=\{1,1,1\}$ is the global atmospheric light.

\subsubsection{Learning to Dehaze from Real-world Hazy Distribution}\label{sec_haze_removal}

\begin{figure*}[pos=!t]
  \centering
  \includegraphics[width=0.7\linewidth]{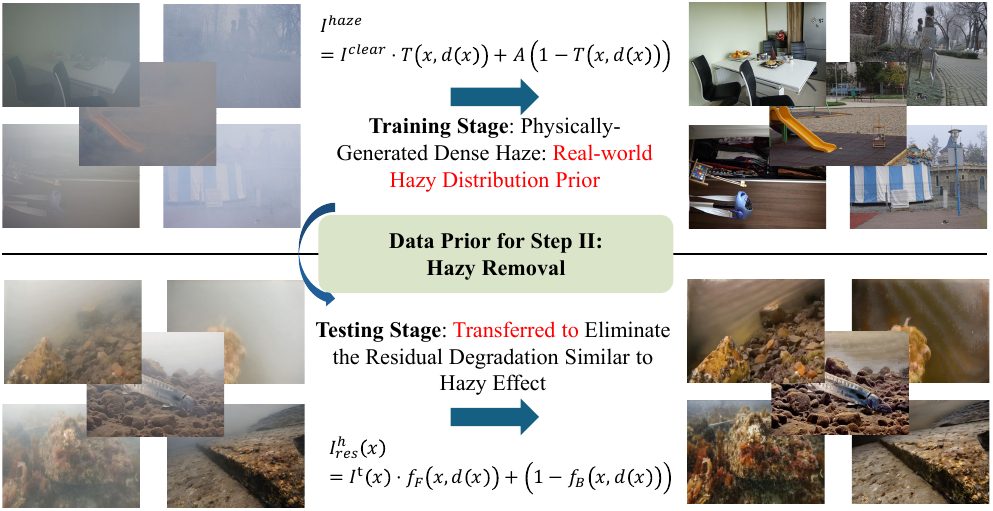}
  \caption{Step II: Hazy Removal based on Real-world Dense Haze Dataset.}
  \label{task2} 
\end{figure*}

Because the residual effect is homogeneous to the in-air hazy effect with white-like atmospheric parameters, another diffusion model is fine-tuned to remove this type of degradation, as illustrated in Fig.\ref{task2}. Unlike color correction training, where the real-world paired dataset with true labels is not available, there is a hazy image dataset named DenseHaze \cite{ancuti2019densehaze} that has both the real dense hazy images and its true labels. The real hazy images are captured in a physically-generated fog environment.

Taking advantage of the real-world densely-hazy images with true labels, the fine-tuned diffusion model learns well about the real-world hazy distribution and is able to dehaze the color-corrected images to a very clear and natural pattern on the foreground part. As demonstrated in the bottom of Fig.\ref{task2}, the hazy effect in the processed images after global color correction has been effectively eliminated, resulting in the expected enhanced images with high contrast. However, the background part of some enhanced images is noisy, as is observed. This reason behind is that although using dense hazy images contributes to a clear and high contrast result, the dense hazy effect diminishes the distinction between the foreground and background, which in turn leads to an unreasonable enhancement of the background contrast. An ideal enhanced image should have a high contrast pattern in the foreground area, while the background area is always expected to be smooth and not so noisy. Therefore, a further process is required to address the noisy background problem. 

\subsubsection{Semantic Segmentation for Background Noise Suppression with Automatically Generated Prompts}\label{sec_bg_noise_suppression}

\begin{figure*}[pos=!t]
  \centering
  \includegraphics[width=0.7\linewidth]{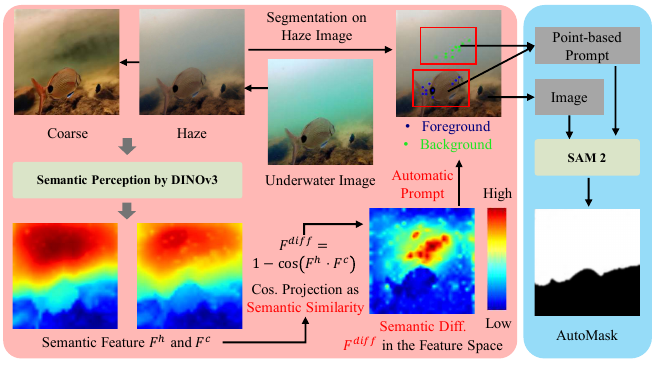}
  \caption{The process of Automatic Foreground-Background Masking based on Semantic foundation models. The semantic features are colorized after dimensionality reduction by Principal Component Analysis(PCA), while the semantic difference image is normalized to 0-1 and then colorized.}
  \label{fuseproc} 
\end{figure*}

\begin{figure*}[pos=!t]
  \centering
  \includegraphics[width=0.7\linewidth]{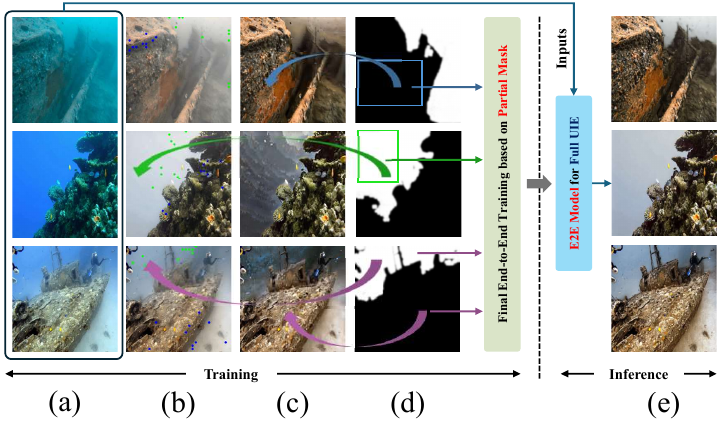}
  \caption{Examples of the final end-to-end training with background noise suppression conducted. Columns (a), (b), (c) are the raw underwater images, enhanced results after global color correction, and enhanced results after haze removal, respectively. Columns (d) and (e) denote the segmentation masks and corresponding results of end-to-end training.}
  \label{fuseres} 
\end{figure*}
Notably, although the output of the color correction task is hazy, the smooth results in the background part meet the requirement mentioned above. This is because the hazy effect is the residual in the enhanced result of global color correction. Such a hazy and homogeneous background with the correct color pattern suppresses the noisy background to improve the visual quality as shown in Fig.\ref{fuseres} (b), (c). Hence, we additionally implement semantic segmentation on the hazy images to extract the expected background part to further refine the enhanced result.

Specifically, the embedded prior of two pre-trained semantic models Segment Anything v2 (SAM2) \cite{ravi2024sam2} and DINOv3 \cite{simeoni2025dinov3} are used to segment the two parts. As illustrated on the right-hand side of Fig.\ref{fuseproc}, SAM2 labels the foreground and background parts according to the input images and the point-based prompt to generate masks. Manually labeling the foreground-background prompt is quite laborious and is not suitable for scalable label generation. Therefore, an automatic prompt generation strategy is designed based on the semantic difference of enhanced images after global color correction ("haze") and dehazed images with noisy background ("coarse"). First, the semantic features of these two images are extracted by DINOv3 \cite{simeoni2025dinov3}. The semantic difference is then defined as the cosine projection:
\begin{equation}
   F^{diff}=1-\cos(F^h\cdot F^c),
\end{equation}
where $F^h,F^c$ are the semantic features of the haze and coarse images, respectively. After that, the Top N pixels with the maximal feature difference are selected as the background point prompts because of the dramatic semantic changes induced by noise. In contrast, another Top N pixels with minimal feature difference are selected as the foreground part prompts because of the stable semantic information. The above process of automatic prompt generation is illustrated in the right hand side of Fig.\ref{fuseproc}. In this way, human intervention is greatly minimized, leading to a significant reduction in data processing time and enabling scalable data labeling, since prompt labeling is the most time-consuming part.

Although the proposed strategy supports automatic prompt generation for masking, the DNIOv3-SAM2 pipeline cannot generate perfect masks in some cases. Therefore, a partial training strategy is also proposed: 1) if the foreground segmentation is correct (no need to fully segmented, but the labeled foreground part is fully correct as shown in the first row of Fig.\ref{fuseres}), then only the foreground part of the dehazed image is used for training; 2) if the background segmentation is correct (also no need to fully segmented but the labeled background part is fully correct as shown in the second row of Fig.\ref{fuseres}), then only the background part of the hazy image is used for training; 3) if both the foreground and background parts are segmented in a reasonable way, then these two parts are randomly used for partial training. 4) if both of these two parts are segmented incorrectly, then the image is discarded. In summary, the final End-to-End model learns a region-wise enhancement strategy to properly suppress the noise of the background but to increase the clarity of the foreground, as shown in Fig.\ref{fuseres} (e).

\subsection{Diffusion-based Formulation for Image Enhancement}

\begin{figure}[pos=!t]
  \centering
  \includegraphics[width=1\linewidth]{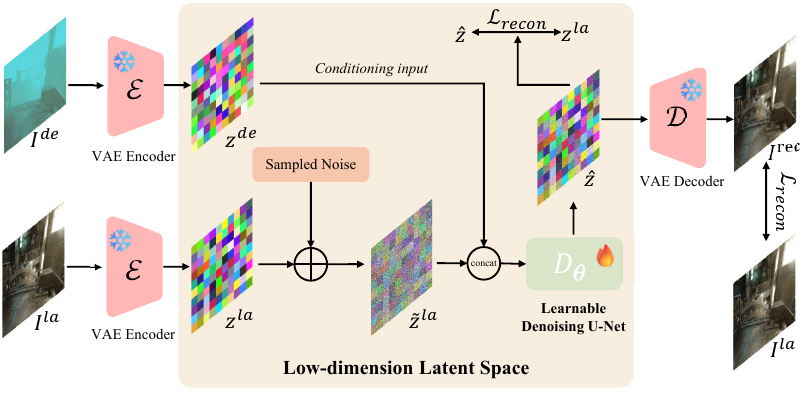}
  \caption{The pipeline of fine-tuning diffusion model in the latent space.}
  \label{ldm} 
\end{figure}

As mentioned in Section \ref{dm_relatedwork}, pre-trained diffusion models demonstrate previously unattainable capabilities in the field of high-quality image generation. Therefore, the models used in the global color correction, hazy removal, and final end-to-end enhancement are all from an open-source pre-trained model \footnote{https://huggingface.co/stabilityai/stable-diffusion-2}, and the whole process is depicted in Fig.\ref{ldm}. 

In this work, these three steps for image enhancement are uniformly reformulated as the task of conditional denoising diffusion generation, and the pre-trained diffusion model providing an extensive embedded real-world scene prior is used in all these three steps for initialization to accelerate the training process. The diffusion model learns the probabilistic distribution of the data $\mathbf{x}_0 \sim p(\mathbf{x})$ through a forward diffusion process and a reverse denoising process followed, where $\mathbf{x}_0, \mathbf{x}$ are the expected enhanced images and the corresponding degraded images, respectively. In the forward diffusion process, the model first adds a certain noise pattern to the data to destroy its structure, then it learns to reverse the process by denoising the initial noisy input to reconstruct the original data under conditioning input which are degraded images.  

To ensure computational efficiency and suitability for high-resolution image generation, the denoising process is implemented in a low-dimensional latent space as adopted in \cite{garcia2025e2eft,hu2025pfusie}. Instead of directly working on the pixel space, the latent space is enclosed by a Variational AutoEncoder (VAE) that is also pre-trained independently. The VAE consists of two components: an encoder $\mathcal{E}$ and a decoder $\mathcal{D}$. The input $\mathbf{x}$, which is a degraded or target image in this proposal, is transformed by VAE: $\mathbf{z}^{(\mathbf{x})}=\mathcal{E}(\mathbf{x})$, $\mathbf{\hat x}=\mathcal{D}(\mathbf{z}^{(\mathbf{x})})$, where $\mathbf{z}^{(\mathbf{x})}$, $\mathbf{\hat x}$ are the latent representation and the reconstructed version of the input image, respectively. 

The denoiser, which is a U-net $D_\theta$ with network parameters $\theta$, is fine-tuned in the latent space to perform image enhancement. In the training stage, the degraded image $I_{de}$ and its enhanced paired image as the label $I_{la}$ are encoded by the VAE encoder $\mathcal{E}$ as:
\begin{equation}
    \mathbf{z}_{de}=\mathcal{E}(I_{de})\in\mathbb{R}^{W\times H \times D},\mathbf{z}_{la}=\mathcal{E}(I_{la})\in\mathbb{R}^{W\times H \times D},
\end{equation}
where $H,W,D$ are the height, width and dimension of the latent space. The noise $\epsilon$ following a certain distribution $p$ is then randomly sampled and added to $\mathbf{z}_{la}$ as $\mathbf{\tilde{z}}_{la}$, which is concatenated with the conditioning input $\mathbf{z}_{de}$ as $\mathbf{\tilde{z}}_{cat}\in\mathbb{R}^{W\times H \times 2D}$. $\mathbf{\tilde{z}}_{cat}$ will then be denoised by $D_\theta$ to a latent estimation $\mathbf{\hat z}$, which should ideally align with $\mathbf{z}_{la}$. The loss function is given as:
\begin{equation}
    \mathcal{L}=\mathbb{E}_{\epsilon\sim p}||\mathbf{\hat z}-\mathbf{z}_{la}||^2_2.
\end{equation}
For the diffusion model trained for the final end-to-end enhancement, the loss in the original pixel-space space in a partial training manner is alternatively given as:

\begin{equation}
    \mathcal{L}=\mathbb{D}_{\epsilon\sim p}||(\mathcal{D}(\mathbf{\hat z})-I_{la})\cdot M||^2_2,  
\end{equation}
where $M$ denotes masks with 
\begin{equation}
    M^b(x)=0,M^f(x)=1
\end{equation}
for the foreground partial training and
\begin{equation}
    M^b(x)=1,M^f(x)=0
\end{equation}
for the background partial training, where $x$ is the pixel and the superscripts $f,b$ represent the foreground and background, respectively.

\section{Experiments Results}
\subsection{Implementation Details}

For the global color correction step, the synthetic underwater images are based on the TARTAN \cite{wang2020tartanair} dataset. The TARTAN dataset is a synthetic dataset that contains various indoor and outdoor scenarios. It is collected from a photorealistic simulation platform. The ground truth depth map allows for the UIFM-based underwater image synthesis. For the haze removal step, the DenseHaze dataset \cite{ancuti2019densehaze} is selected for training. It contains 33 pairs of hazy image-ground truths of various outdoor scenes, where the hazy effect is real and is generated by professional haze machines. Therefore, a real-world hazy distribution is available for the model to dehaze images well. 

For quantitative and qualitative comparison, two real-world underwater image datasets LSUI \cite{peng2023ushape} and EUVP \cite{islam2020euvp} are selected. LSUI contains 4279 underwater images with pseudo-labels, while EUVP contains 6665 underwater images with the labels generated by the GAN-based approach. We do not use noisy labels of the LSUI and EUVP for full-reference comparisons such as peak signal-to-noise ratio (PSNR) and structural similarity (SSIM). Instead, we mainly take the Non-Reference Image Quality Assessment (NR-IQA) for the performance evaluation. In Semi-UIR \cite{huang2023semiuir}, Multi-Scale Image Quality (MUSIQ) \cite{ke2021musiq} and From Patches to Pictures (PAQ2PIQ) \cite{ying2020paq2piq} rank first in terms of metric reliability, while Uranker \cite{guo2023uranker} and MUSIQ are first as analyzed in \cite{guo2023uranker}. Therefore, Uranker, MUSIQ, and PAQ2PIQ are selected as NR-IQA indies to quantitatively evaluate the algorithm performance. Because some previous works \cite{huang2023semiuir,cong2023pugan,pucci2025cevae,mei2025dpf,peng2025ssuie} fix the evaluation resolution as 256x256, we follow them to compare the results with such image resolution. The proposal is compared with ten SOTA UIE methods developed in recent two years, including Semi-UIR (2023, CVPR) \cite{huang2023semiuir}, PUGAN (2023, TIP) \cite{cong2023pugan}, CCL (2024, TMM) \cite{liu2024ccl}, HCLR (2024, IJCV) \cite{zhou2024hclr}, WF-Diff (2024, CVPR) \cite{zhao2024wfdiff}, DPF-Net (2025, ISPRS) \cite{mei2025dpf}, CE-VAE (2025, WACV) \cite{pucci2025cevae}, SS-UIE (2025, AAAI) \cite{peng2025ssuie}, UIE-CLIP (2025, ISPRS) \cite{cao2025uieclip}, and PFUSIE (2025, IF) \cite{hu2025pfusie}.

The diffusion models used in the color correction, haze removal, and the final end-to-end UIE task are all fine-tuned from the pre-trained version of Stable Diffusion \footnote{https://huggingface.co/stabilityai/stable-diffusion-2} on 4 RTX3090 GPUs for 80k iterations. The detailed fine-tuning strategy is followed by \cite{garcia2025e2eft} with a fixed timestep and deterministic noise distribution. The learning rate is set to 3e-5, with the Adam optimizer parameters as $\beta_1 = 0.9$ and $\beta_2=0.999$. Due to the GPU memory limitation, the batch size is set to 1 per GPU and the resolution of training images is fixed to 512x512. The gradient accumulation step is set to 20, which means that the effective batch size is 80. For the TARTAN dataset, we set the surface-to-object depth randomly sampled from $[0, 5]$ followed by \cite{li2020uwcnn}, and the object-to-sensor distance is set to 0/5 for the foreground/background area, respectively. In addition, the water type $\mathrm{Nrer}=\{\mathrm{Nrer}_R,\mathrm{Nrer}_G,\mathrm{Nrer}_B\}$ is set in the range of:
\begin{equation}
\sum_{\lambda}\mathrm{Nrer}_{\lambda} > 1.3,
\end{equation}
where $\max_{\lambda}\mathrm{Nrer}_{\lambda} > 0.6,\mathrm{Nrer}_{\lambda} \in (0,1)$ followed by \cite{hu2025pfusie}. For the DenseHaze dataset, because there are only 33 pairs for training, a data augmentation strategy is implemented: 1) The original images are randomly cropped to 768x768; 2) Then the cropped images are randomly rotated with a maximum rotation angle of 30 degrees; 3) The augmented images are resized to 512x512 for training.

In the background noisy superssion stage, the SAM2 API \footnote{https://github.com/facebookresearch/sam2} is utilized for prompted-based foreground-background segmentation. To fine-tune the final end-to-end underwater image enhancement diffusion model, The following detailed steps are implemented:

\begin{itemize}
    \item We tested fine-tuned color correction and haze removal diffusion models on four other underwater datasets UIEB \cite{li2019uieb}, UVE38K \cite{qi2021uve38k}, UVEB \cite{xie2024uveb} and DRUVA \cite{varghese2023druva}. The UVE38K, DRUVA, and UVEB datasets are three underwater video datasets that include various scenarios and water conditions. The UIEB dataset is a full reference underwater image dataset that is widely used for most supervised learning-based methods, although the labels are also noisy. For each scene in the three video datasets, several images with a proper sampling step are selected to remove color bias and the hazy effect.
    \item The proposed semantic-based pipeline is used to automatically generate masks. This is followed by a very lightweight manual selection step, where the generated masks are labeled for partial training. Each mask is categorized as one of the following: "foreground training," "background training", "full training", or "discarded.". "full training" means that both parts are available for partial training. 
    \item The end-to-end UIE diffusion model is fine-tuned by the collected dataset P2UIE\_1473 (800 from UIEB and 673 from three video datasets) with mask guidance introduced in \ref{sec_bg_noise_suppression}. To show that the performance gain can be attributed to the proposal rather than the increasing number of training images (most of the comparison methods are trained by UIEB), we only use the UIEB subset of P2UIE\_1473 for training for a fair comparison, but we tested the entire P2UIE dataset. Notice that it causes no data leakage because we use non-reference metrics that are not calculated based on the ground-truth for performance evaluation.
    
\end{itemize}

\subsection{Qualitative and Quantitative Results Comparison}

    \begin{table*}[pos=!t]
    \centering
    \scalebox{0.8}{
    \begin{tabular}{c|ccc|ccc|ccc|c}
    \multirow{2}{*}{Method} & \multicolumn{3}{|c|}{EUVP\cite{islam2020euvp}} & \multicolumn{3}{c|}{LSUI\cite{peng2023ushape}} & \multicolumn{3}{c|}{P2UIE\_1473} & \multirow{2}{*}{Avg. Rank $\downarrow$}\\
    \cline{2-10}
     & {\color{green}PAQ2PIQ}$\uparrow$ & {\color{blue}MUSIQ}$\uparrow$ & {\color{red}Uranker}$\uparrow$ & {\color{green}PAQ2PIQ}$\uparrow$ & {\color{blue}MUSIQ}$\uparrow$ & {\color{red}Uranker}$\uparrow$ & {\color{green}PAQ2PIQ}$\uparrow$ & {\color{blue}MUSIQ}$\uparrow$ & {\color{red}Uranker}$\uparrow$ \\
    \hline
    Semi-UIR\cite{huang2023semiuir} & \underline{74.909} & 50.410 & \underline{2.472} & \underline{72.565} & 43.268 & \underline{2.365} & 73.018 & 46.358 & 1.993 & \underline{3.33} \\
    PUGAN\cite{cong2023pugan} & 74.323 & 46.751 & 2.383 &71.734 & 39.171 & 2.264 & 73.192 & 42.197 & \underline{2.091} & 6.00\\
    CCL\cite{liu2024ccl} &  74.393 & 48.638 & 2.051 & 71.876 & 40.662 & 1.954 & \textbf{73.659} & 47.703 & 1.722 & 5.78\\
    HCLR\cite{zhou2024hclr} & 74.094 & 48.330 & 2.235 & 71.227 & 39.800 & 2.112 & 72.774 & 45.002 & 1.992 & 6.89 \\
    WF-Diff\cite{zhao2024wfdiff} & 73.556 & 51.483 & 2.057 & 70.594 & \underline{44.922} & 1.805 & 70.062 & 46.083 & 1.518 & 8.11\\
    CE-VAE\cite{pucci2025cevae} & 74.215 & 49.498 & 1.897 & 71.816 & 43.919 & 1.678 & 71.001 & 42.243 & 1.562 & 8.00 \\
    SS-UIE\cite{peng2025ssuie} & 73.672 & 49.750 & 1.931 & 71.037 & 42.538 & 1.812 & 71.523 & 46.624 & 1.765 & 7.44 \\
    DPF-Net\cite{mei2025dpf} & 73.633 & 47.976 & 2.196 & 70.734 & 40.314 & 2.040 & 71.631 & 47.588 & 1.730 & 7.78 \\
    PFUSIE\cite{hu2025pfusie} & 73.067 & \underline{51.492} & 2.378 & 71.891 & 44.165 & 2.169 & 72.968 & \underline{48.958} & 1.841 & 4.44 \\
    UIE-CLIP\cite{cao2025uieclip} & 74.110 & 48.747 & 2.164 & 71.014 & 40.568 & 2.012 & 72.539 & 45.610 & 1.909 & 7.11 \\
    
    \hline
     \textbf{Ours} & \textbf{74.921} & \textbf{54.598} & \textbf{2.553} &\textbf{73.417} & \textbf{48.715} & \textbf{2.502} & \underline{73.315} & \textbf{52.288} & \textbf{2.409} & \textbf{1.11}\\
    \end{tabular}}
    \caption{Comparison of ten different methods on three datasets and three metrics (the higher is better). For each metric of a given dataset, the best result is in \textbf{bold}, the result ranks 2th is with \underline{underline}.}
    \label{metric_comp}
    \end{table*}

\begin{figure*}[pos=!t]
  \centering
  \includegraphics[width=1.0\linewidth]{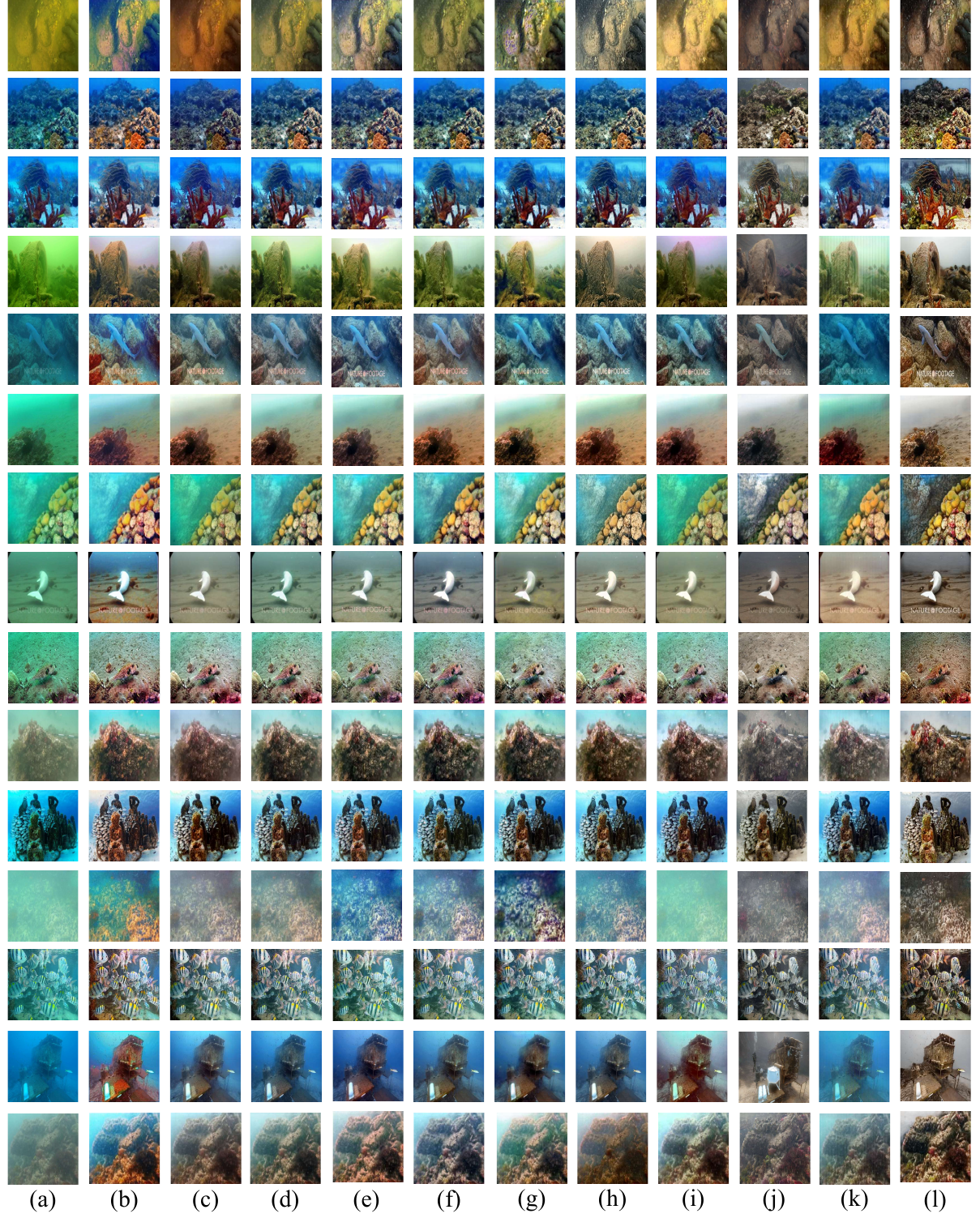}
  \caption{Quantitative results on three datasets. Row 1-4, 5-9, 10-15 are sampled from dataset EUVP, LSUI and our self-collected P2UIE\_1473, respectively. Column (a) represents the raw underwater images, while columns (b)-(k) are the enhanced results of previous works CCL \cite{liu2024ccl}, CE-VAE \cite{pucci2025cevae}, DPF \cite{mei2025dpf}, UIE-CLIP \cite{cao2025uieclip}, HCLR \cite{zhou2024hclr}, PUGAN \cite{cong2023pugan}, Semi-UIR \cite{huang2023semiuir}, SSUIE \cite{peng2025ssuie}, PFUSIE \cite{hu2025pfusie}, WFDiff \cite{zhao2024wfdiff}. Column (l) illustrates the enhanced results of the proposal.}
  \label{res_comp1} 
\end{figure*}

Tab.\ref{metric_comp} shows the quantitative results of different methods with respect to three test datasets and three metrics. The average rank is calculated by 
\begin{equation}
\sum rank^{metric}_{dataset}/(N_{dataset}\cdot N_{metic}),
\end{equation}
where $rank_{dataset}^{metric}$ is the rank of a method among all methods of a dataset-metric pair. The proposal is far superior to previous works, ranking first for 8 of 9 metrics. The second and third ranking methods are Semi-UIR and PFUSIE, but the ranking value obviously drops. The reason behind this is that our method is based on transfer learning and cross-domain supervision. We do not use the heavily biased label of underwater images for end-to-end supervision or simply use fully synthetic underwater images for training, both causing non-ignorable synthetic-real discrepancy. Instead, the UIE is decomposed into global color correction and haze removal, where the diffusion prior and the real-world hazy distribution are leveraged to address these two problems, respectively. Furthermore, the foreground-background understanding embedded in the semantic model also helps to generate enhanced images with noise-suppressed background to have better visual quality. Although there is a lack of true labels for underwater images, this physics-based task decomposition and transfer learning strategy enables the feasibility of enhancing underwater images well. In contrast, while the mentioned benchmark methods have made much progress, they mainly focus on network design and training strategy for label-alignment, paying little attention to the issue of unreliable labels. Meanwhile, the qualitative comparison result is illustrated in Fig.\ref{res_comp1}, where the advantage of our proposal is more obvious, showing a correct color pattern, a high contrast foreground and a smooth background.

\begin{figure*}[pos=!t]
  \centering
  \includegraphics[width=1.0\linewidth]{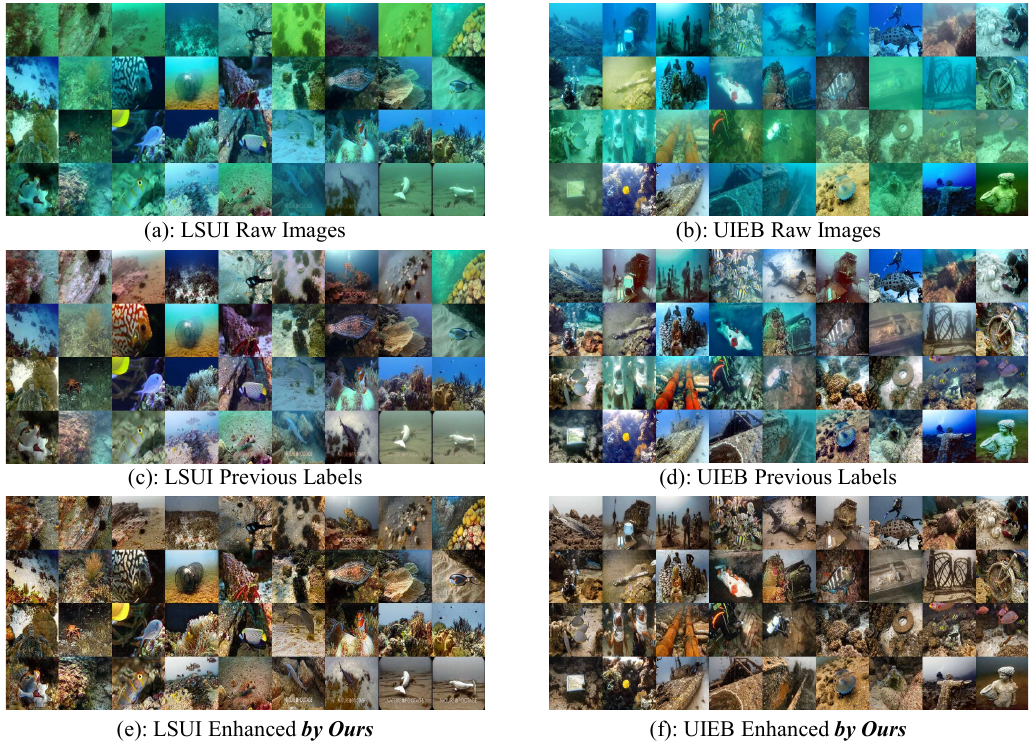}
  \caption{An overview of the raw images ((a),(b)), previous noisy labels ((c),(d)), and the enhanced results of the proposed P2UIE ((e),(f)). (a), (c), (e) are from the dataset LSUI, while (b), (d), (f) are from the dataset UIEB.}
  \label{pre_gt_comp} 
\end{figure*}

When physical settings are configured at the stage of image synthesis, the water parameters vary in a continuous range, including different color bias conditions. In addition, the depth is also set as a range containing different conditions of visual clarity, from pure clear to almost invisible. Moreover, The real-world heavily hazy dataset also provides the knowledge to effectively remove the depth-related degradation. Therefore, the model has witnessed various water effects from slight degradation to heavy degradation, and is able to significantly improve the quality of underwater images as shown in Fig.\ref{pre_gt_comp}, which is the overall visual result compared to the previous noisy labels and raw underwater images in the two paired datasets LSUI and UIEB. The performance of P2UIE is better than that of the models supervised by the previous labels and, more importantly, more stable among different images under various degradation conditions.

\begin{figure}[pos=!t]
  \centering
  \includegraphics[width=1.0\linewidth]{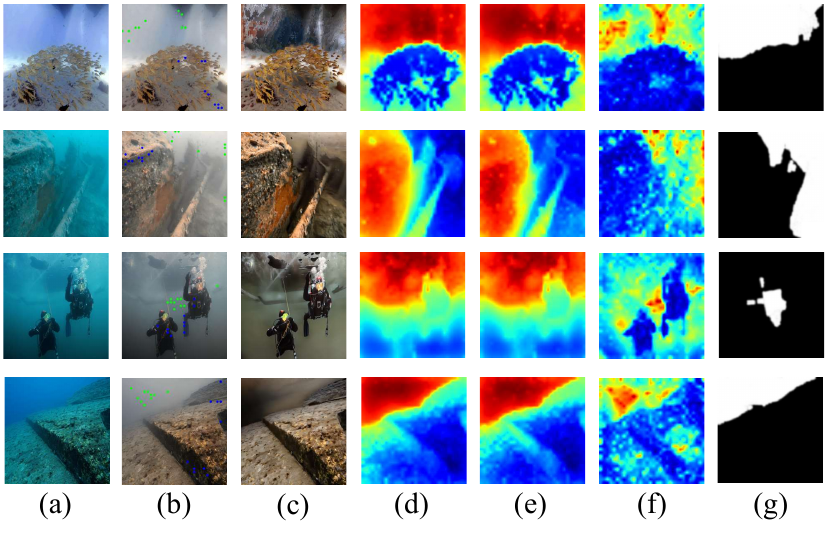}
  \caption{For each row, columns (a) to (g) represent the underwater image (a), the color corrected image (b), the dehazed image (c), the DINO feature of the color corrected image (d), the DINO feature of the dehazed image (e), the feature difference of (d) and (e), and the generated mask based on SAM2 from automatic prompt, respectively.}
  \label{feat_diff_vis} 
\end{figure}

Fig.\ref{feat_diff_vis} shows additional results of semantic differences in the feature space for automatic prompt generation. For rows 1, 2, the generated masks only support partial training (background and foreground, respectively). For rows 4, both parts are segmented correctly. Therefore, the generated mask supports training for either part. However, for row 3, although the prompt automatically generated from semantic difference is reasonable, the mask is not suitable for training and is discarded. In our experiment, the numbers of generated masks that 1) only support partial training 2) training for both parts 3) not suitable for training are 539, 197, 64, respectively. By the proposed partial training, the model learns to process the two areas in a different way to simultaneously 1) emphasize the foreground part while 2) smooth the background part to suppress possible noise. 

\begin{figure*}[pos=!t]
  \centering
  \includegraphics[width=1.0\linewidth]{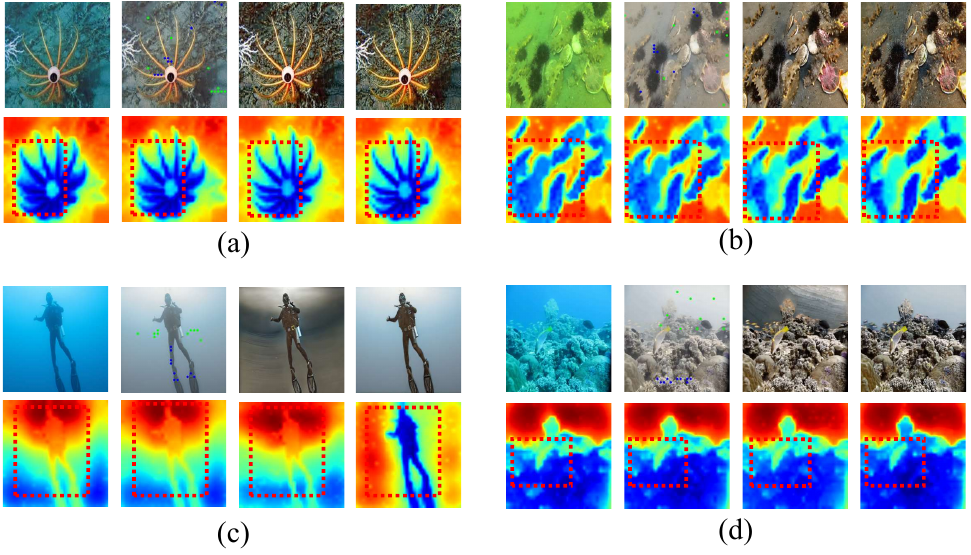}
  \caption{Illustration of the intermediate results. For each subfigure, the first row shows the underwater image, the color corrected image, the dehazed image, and the final enhanced image, respectively; while the second row represents the corresponding semantic feature extracted from DINO model and visualized by PCA.}
  \label{inter_fes_ana} 
\end{figure*}

Fig.\ref{inter_fes_ana} illustrates details about the intermediate results of the proposed framework, including the local enhancement results and the corresponding semantic features. Semantic features are visualized by selecting the first three main components of the PCA. Semantic clustering in the feature space is observed to become more precise in the proposed framework process (red rectangles), and the final enhanced results demonstrate the best performance of semantic perception. This phenomenon is very obvious in the left-bottom scene, where the diver feature in the final result is much more different from the background than in the previous raw image or in the intermediate results. There are two main reasons: first, the background noise is suppressed and less informative; second, the texture of the foreground part is forced to be emphasized and greatly improved, which is reversely more informative. Consequently, the final enhanced results show high visual quality in both the pixel and semantic feature space.

\begin{figure}[pos=!t]
  \centering
  \includegraphics[width=1\linewidth]{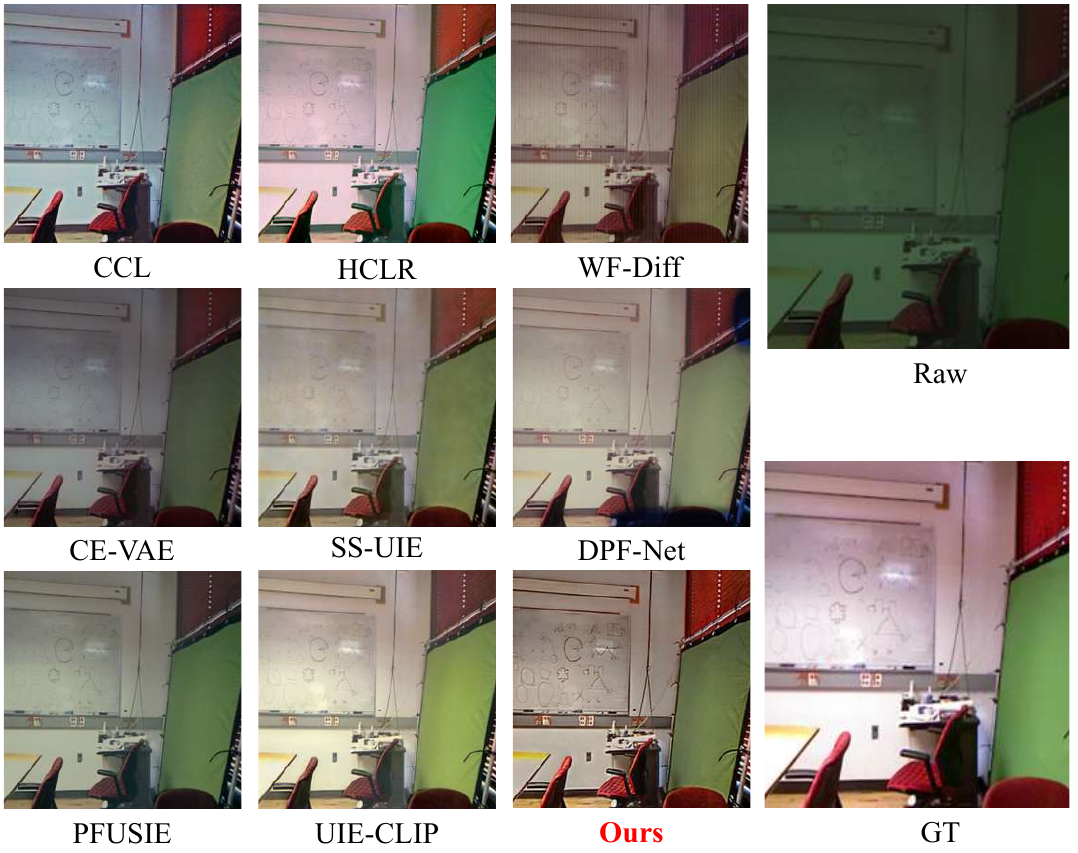}
  \caption{Quantitative performance comparison on the synthetic dataset UWCNN \cite{li2020uwcnn}.}
  \label{uwcnn_visual_ana} 
\end{figure}

The full-reference performance evaluation is also conducted on two synthetic datasets: Replica \cite{straub2019replica} and UWCNN \cite{li2020uwcnn}. Since SemiUIR \cite{huang2023semiuir} and PUGAN \cite{cong2023pugan} are trained using UWCNN \cite{li2020uwcnn}, we exclude their full-reference metrics on the UWCNN dataset. To enable a comprehensive comparison including SemiUIR and PUGAN, we also follow UWCNN to synthesize an underwater dataset on Replica \cite{straub2019replica}. As shown in Tab.\ref{psnrssim_syn}, the proposed P2UIE achieves competitive average performance. Moreover, Fig.\ref{uwcnn_visual_ana} presents a visual comparison, demonstrating that P2UIE achieves better visual clarity among the evaluated UIE methods.

\begin{table}[pos=!t]
\centering
\scalebox{0.9}{
\begin{tabular}{lcccc}
\hline
Method & \multicolumn{2}{c}{UWCNN} & \multicolumn{2}{c}{Replica} \\
\cline{2-5}
       & PSNR & SSIM & PSNR & SSIM \\
\hline
SemiUIR\cite{huang2023semiuir} & - & - & {\color{green}17.3238} & {\color{green}0.7902} \\
PUGAN\cite{cong2023pugan}   & - & - & 17.0560 & 0.7864 \\
CCL\cite{liu2024ccl}     & {\color{green}14.1947} & 0.6571 & 13.6678 & 0.7271 \\
HCLR\cite{zhou2024hclr}    & 13.8729 & {\color{blue}0.6788} & 15.1753 & 0.7816 \\
WF-Diff\cite{zhao2024wfdiff} & 13.8815 & 0.6005 & 14.6689 & 0.6366 \\
CE-VAE\cite{pucci2025cevae}  & 13.9209 & 0.6222 & 16.0331 & 0.7590 \\
SS-UIE\cite{peng2025ssuie}  & 13.7279 & 0.6447 & {\color{blue}17.3238} & 0.7631 \\
DPF-Net\cite{mei2025dpf} & {\color{blue}14.1770} & 0.6554 & 16.9647 & 0.7628 \\
PFUSIE\cite{hu2025pfusie}  & 13.8916 & 0.6682 & 15.5196 & 0.7523 \\
UIE-CLIP\cite{cao2025uieclip} & 14.0991 & {\color{red}0.6909} & 16.3065 & {\color{red}0.8009} \\
Ours    & {\color{red}15.2018} & {\color{green}0.6704} & {\color{red}17.5133} & {\color{blue}0.7937} \\
\hline
\end{tabular}}
\caption{PSNR and SSIM comparison on UWCNN \cite{li2020uwcnn} and Replica 
\label{psnrssim_syn}
\cite{straub2019replica} datasets. Colors {\color{red}red}, {\color{green}green}, and {\color{blue}blue} represents the first, second, and third best performance for each metric and dataset.}
\end{table}

\subsection{Ablation Study and Discussions}

To evaluate the effectiveness of the proposed design, three different ablation experiments were conducted. First, we directly fine-tuned a diffusion model using dataset UIEB for end-to-end UIE task, which means the heavily biased labels are utilized for supervised learning. The results is denoted as \textbf{UIEB}. Second, we replaced the real-world hazy distribution with synthetic hazy images for the haze removal step(the synthetic process is followed by \cite{liu2021haze4k} in the TARTAN dataset \cite{wang2020tartanair}. The result is denoted as \textbf{TARTAN}. Third, we remove the step of background noise suppression. The result is denoted as \textbf{NoSeg} , which means "no segmentation". For each ablation experiment, the other settings are kept the same. For example, in \textbf{TARTAN}, the background noise suppression is preserved, then the fused enhanced images are used for the final end-to-end fine-tuning of a new diffusion model that directly recovers underwater images. The quantitative comparison result with respect to three metrics is shown in Tab.\ref{tduf_ablation}. 
It is observed that both foreground-background segmentation and real-world hazy distribution contribute to performance improvement. In addition, unreliable labels actually impair the overall enhancement quality because the color bias and low contrast effects are not well addressed in the labels.

\begin{table*}[pos=!t]
\centering
\scalebox{0.8}{
\begin{tabular}{c|ccc|ccc|ccc|c}
\multirow{2}{*}{Method} & \multicolumn{3}{|c|}{EUVP\cite{islam2020euvp}} & \multicolumn{3}{c|}{LSUI\cite{peng2023ushape}} & \multicolumn{3}{c|}{P2UIE\_1473} & \multirow{2}{*}{Avg. Rank $\downarrow$}\\
\cline{2-10}
 & {\color{green}PAQ2PIQ}$\uparrow$ & {\color{blue}MUSIQ}$\uparrow$ & {\color{red}Uranker}$\uparrow$ & {\color{green}PAQ2PIQ}$\uparrow$ & {\color{blue}MUSIQ}$\uparrow$ & {\color{red}Uranker}$\uparrow$ & {\color{green}PAQ2PIQ}$\uparrow$ & {\color{blue}MUSIQ}$\uparrow$ & {\color{red}Uranker}$\uparrow$ \\
\hline
NoSeg & 73.421 & 49.753 & \textbf{2.743} & 72.048 & 43.999 & \textbf{2.685} & 72.291 & \underline{50.483} & \underline{2.396} & \underline{2.44} \\
UIEB-Labels & \underline{74.694} & 49.464 & 2.415 &71.962 & 41.202 & 2.217 & \underline{73.207} & 47.574 & 2.067 & 3.11\\
TARTAN-Syn & 74.097 & \underline{50.945} & 2.083 & \underline{72.078} & \underline{44.229} & 2.063 & 71.369 & 47.276 & 1.704 & 3.22\\
\hline
\textbf{FULL} & \textbf{74.921} & \textbf{54.598} & \underline{2.553} &\textbf{73.417} & \textbf{48.715} & \underline{2.502} & \textbf{73.315} & \textbf{52.288} & \textbf{2.409} & \textbf{1.22}\\
\end{tabular}}
\caption{Ablation study on three datasets and three metrics (the higher is better). For each metric of a given dataset, the best result is in \textbf{bold}, the result ranks 2th is with \underline{underline}.}
\label{tduf_ablation}
\end{table*}

We also provide a quantitative comparison on the average time it takes to generate End-to-End training masks in Section \ref{sec_bg_noise_suppression} using the proposed automatic prompt labeling strategy and fully manual labeling. As shown in Tab \ref{prompt_gen_ana}, the manual prompt labeling step takes more than 15 seconds per image, which is the most time-consuming process and prevents the framework from collecting scalable datasets. In contrast, the total time drops to about 86.9\% when replaced by automatic prompt labeling. Although minimal human interaction has to be involved for partial mask labeling, the current time consumption is relatively reasonable compared to 16 seconds per image. The quantitative performance is given in Tab.\ref{auto_mask_ablation}, showing a very slight but justifiable decrease as a compromise of a more automatic pipeline. This is because the partial but correct masks already allow network know to infer which regions should be foreground and which can be smoothed into the background.

\begin{table}[pos=!t]
\centering
\scalebox{1}{
\begin{tabular}{c|cc}
Avg. Time Cost (s)  & Manual Prompt & Auto Prompt \\
\hline
Prompt Generation & 15.3 & 0.7 \\
Partial Mask Labeling & 1.5 & 1.5 \\
Total & 16.8 & 2.2 \\
\end{tabular}}
\caption{The average time cost for generating a mask for the final end-to-end training. "Manual Prompt" means the prompt is fully labeled by human, while "Auto Prompt" means the prompt generation step is automatically processed by the proposed semantic difference-based segmentation.}
\label{prompt_gen_ana}
\end{table}

\begin{table*}[pos=!t]
\centering
\scalebox{0.8}{
\begin{tabular}{c|ccc|ccc|ccc|c}
\multirow{2}{*}{Method} & \multicolumn{3}{|c|}{EUVP\cite{islam2020euvp}} & \multicolumn{3}{c|}{LSUI\cite{peng2023ushape}} & \multicolumn{3}{c|}{P2UIE\_1473} & \multirow{2}{*}{Avg. Rank $\downarrow$}\\
\cline{2-10}
 & {\color{green}PAQ2PIQ}$\uparrow$ & {\color{blue}MUSIQ}$\uparrow$ & {\color{red}Uranker}$\uparrow$ & {\color{green}PAQ2PIQ}$\uparrow$ & {\color{blue}MUSIQ}$\uparrow$ & {\color{red}Uranker}$\uparrow$ & {\color{green}PAQ2PIQ}$\uparrow$ & {\color{blue}MUSIQ}$\uparrow$ & {\color{red}Uranker}$\uparrow$ \\
\hline
Manual & 74.918 & 54.574 & 2.538 & \textbf{73.700} & \textbf{49.290} & \textbf{2.535} & \textbf{73.784} & 52.262 & \textbf{2.415} & 1\\
Auto. & \textbf{74.921} & \textbf{54.598} & \textbf{2.553} & 73.417 & 48.715 & 2.502 & 73.315 & \textbf{52.288} & 2.409 & 1.11\\
\end{tabular}}
\caption{Quantitative performance based on automatic vs. manual masking. The Avg. Rank means the average metric performances compared with previous works in Tab.\ref{metric_comp}.}
\label{auto_mask_ablation}
\end{table*}

\begin{figure}[pos=!t]
  \centering
  \includegraphics[width=1\linewidth]{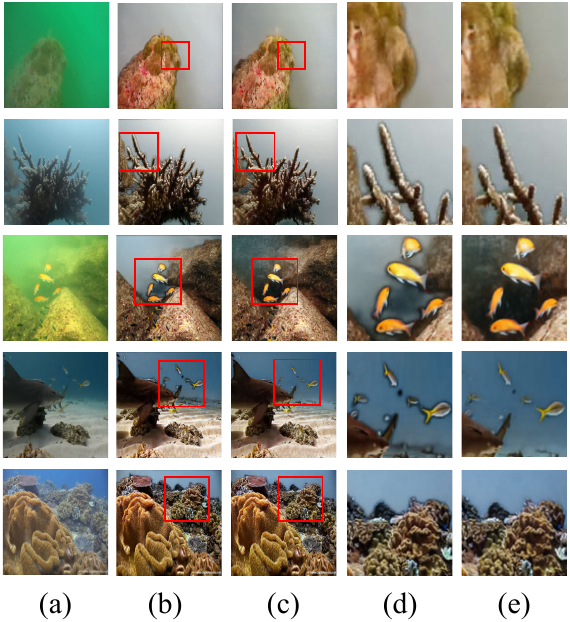}
  \caption{Ablation study of partial training strategy. For each row, columns (a), (b), (c) are the underwater image, the enhanced image with/without partial training strategy, respectively. columns (d) and (e) are the zoom-in details from (b) and (c) to compare the cut-out effect on the edges.}
  \label{stitch_ana} 
\end{figure}

The effectiveness of the proposed partial training strategy instead of directly stitching the dehazed foreground part and smooth background part according to masks is also validated in Fig.\ref{stitch_ana}. In the end-to-end training step, if the images are directly stitched by the segmentation results fusing the foreground part of dehazed images and the smooth background part of the color-corrected images, the cut-off effect on the edges is apparent as shown in Fig.\ref{stitch_ana} (d). This is due to some of the inaccurate segmentation results in the edge area. By the proposed partial training strategy where most of the images are only used for foreground or background training, the model avoids being misguided and only learns the smooth transient pattern from images with "full training" label where the visual change in the edge area is not abrupt.

\begin{figure}[pos=!t]
  \centering
  \includegraphics[width=1.0\linewidth]{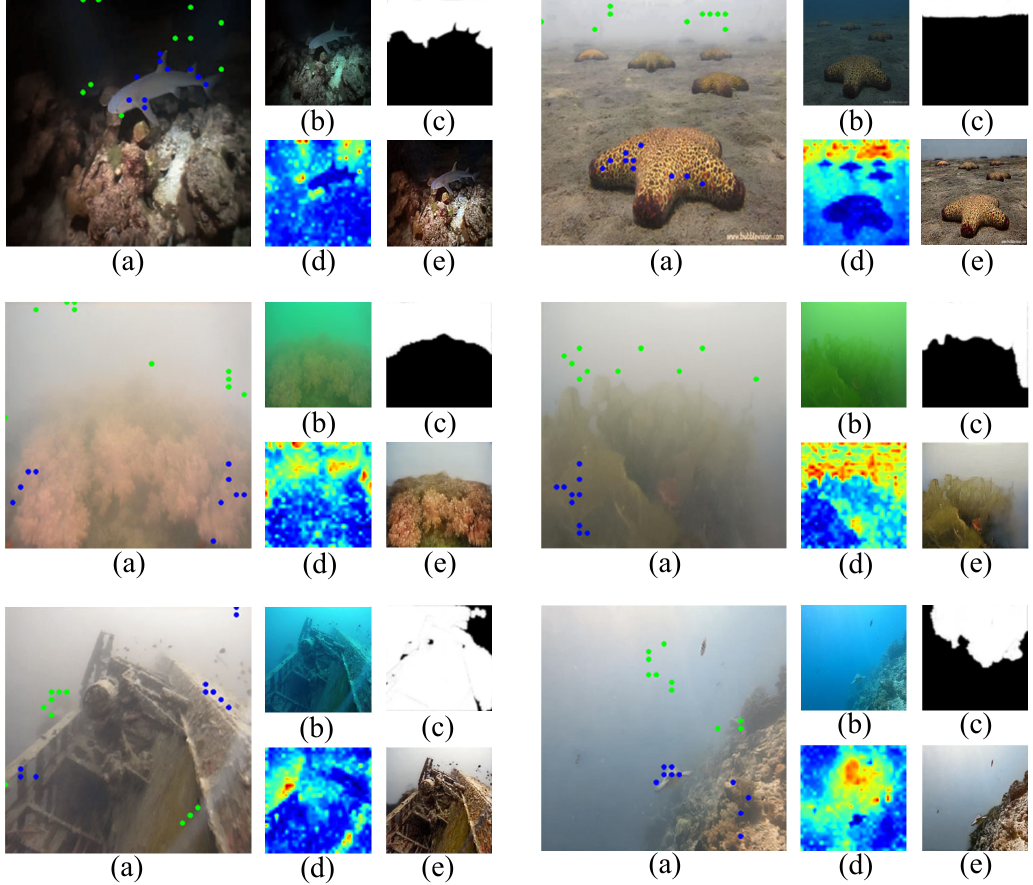}
  \caption{Performance analysis of Dino-based automatic prompt generation strategy. For each scene, subfigure (a)-(e) represent the color corrected image, the raw underwater image, the segmentation image, the feature difference between the color corrected image and haze removed image, and the final result, respectively. The blue and green points in (a) are the foreground/background prompts extracted based on (d) for automatic segmentation.}
  \label{dino_ana} 
\end{figure}

Fig.\ref{dino_ana} illustrates a more detailed performance analysis of the DINO-based automatic mask generation strategy. Note that the extreme color deviation and the extreme low-light condition have already improved in the previous global color-correction step, as shown in the scenes in the first two rows of Fig.\ref{dino_ana} (subfigure (b) to subfigure (a)). The DINOv3 prompt extraction is processed in the color-corrected images. Therefore, it only needs to be and is indeed robust in extreme turbid conditions: 1) qualitatively, the automatically extracted foreground/background prompts in subfigure (a) are reasonable to generate effective segmentation (subfigure (c)); 2) quantitatively, we also calculated the number and proportion of 1) samples that can be used for both training and testing due to effective segmentation and 2) those with useless segmentation results can only be used for testing, which is shown in Tab.\ref{train_test_statistics}. Up to 93.3\% of the samples have valid automatic segmentation based on the correct DINO prompt, which are then used for the final mask-based background noise suppression. There are still some cases where the generated prompts are completely wrong or the segmentation model is not stable on a very small number of problematic points, which makes the failure of segmentation and the corresponding underwater images could only be used for testing. However, the final E2E model trained with effective labels still generates reasonable results (subfigure (e) of the two scenes in the third row of Fig.\ref{dino_ana}).

\begin{table}[pos=!t]
\centering
\begin{tabular}{c|c}
\textbf{Type} & \textbf{Samples} \\
\hline
Train\&Test  & 1374 (93.3\%)\\
Only Test   & 99 (6.7\%) \\
Full dataset  & 1473 \\
\end{tabular}
\caption{Sample counts for train\&test and test-only.}
\label{train_test_statistics}
\end{table}

\begin{figure}[pos=!t]
  \centering
  \includegraphics[width=1.0\linewidth]{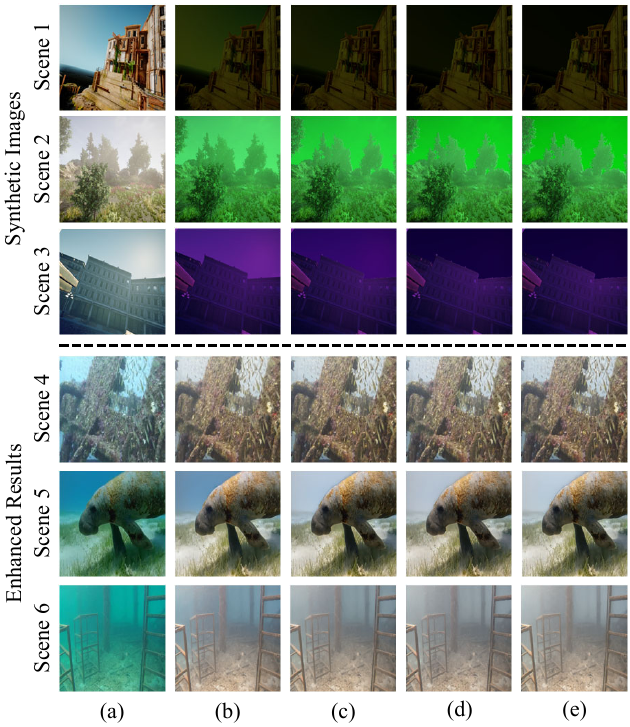}
  \caption{Illustration of the synthetic images and enhanced results for global color correction under different settings of $d_2$. For each scene, the subfigure (a) is the ground truth (for synthetic images)/ raw underwater image (for enhanced results), where (b)-(e) are the synthetic/enhanced images under $d_2\in[0,5], d_2=2.5, d_2=5.0, d_2=7.5$, respectively.}
  \label{horizontal_depth_ana} 
\end{figure}

In the global color correction stage (Section \ref{sec_color_correction}), image synthesis is defined as:
\begin{equation}
I_{\lambda}^{c}=I_{\lambda}^{t}\cdot \mathrm{Nrer}_{\lambda}^{{D(x)+d_F}}+B_\lambda(1-\mathrm{Nrer}_{\lambda}^{d_B}).
\end{equation}
We fix $d_B$ as $d_1=0$ for the foreground part to leave the depth-related residual effect to be decoupled and solved in the haze removal stage (Section \ref{sec_haze_removal}), which is already proven to be effective. Furthermore, the reason to set the background part as 5 instead of 0-5 is that we want to enlarge the pattern difference between the foreground and background area so that the model could have a different color correction strategy on these two areas, as shown in scene 4-6, subfigure (d) in Fig.\ref{horizontal_depth_ana}: 1) For the foreground area, we want the model to correct the global color-bias and simultaneously preserve the original color information under water-free condition; 2) For the background area, we rather want the model to remove the color information to a general white-like hazy background for noise suppression (Section \ref{sec_bg_noise_suppression}) to obtain the final fusion results. 

In contrast, if we set $d_2$ from a fixed value of 5 to a continuous range $(0,5)$, the pattern difference between these two areas weakens, resulting in unsatisfied background results. Although the images only have foreground area have little difference (scene 4 in Fig.\ref{horizontal_depth_ana}), those underwater images with background area still have obvious residual color bias as shown in scenes 4-6, subfigure (b) in Fig. \ref{horizontal_depth_ana}. This is because the continuous change from 0-5 misguides the model to clearly distinguish different regions for region-wise color processing strategies, where it may treat the background part as foreground part to preserve some color bias. This is similar to $d_2=2.5$, where the guidance is not sufficient. Note that the unit of $d$ should be understood as parameters used to control the degree of degradation rather than representing actual meters because when setting $d_2\geq5$ for underwater image synthesis, the background part is already saturated in most cases, as shown in scenes 1-3, subfigure (d) in Fig. \ref{horizontal_depth_ana}. This is consistent with the real-world cases as shown in scenes 4-6, subfigures (d)-(e) in Fig.\ref{horizontal_depth_ana}, where increasing $d_2$ from 5 to 7.5 brings almost no visual difference in the background part at all. We also conducted a quantitative experiment to further support this qualitative result by calculating the metric values under $d_2=5$ and $d_2=7.5$ as shown in Tab.\ref{horizontal_depth_metric_comp}. The close metric performance and identical average rank confirm that a $d_2\geq5$ is already sufficient.

\begin{table*}[pos=!t]
\centering
\scalebox{0.8}{
\begin{tabular}{c|ccc|ccc|ccc|c}
\multirow{2}{*}{Method} & \multicolumn{3}{|c|}{EUVP\cite{islam2020euvp}} & \multicolumn{3}{c|}{LSUI\cite{peng2023ushape}} & \multicolumn{3}{c|}{P2UIE\_1473} & \multirow{2}{*}{Avg. Rank $\downarrow$}\\
\cline{2-10}
 & {\color{green}PAQ2PIQ}$\uparrow$ & {\color{blue}MUSIQ}$\uparrow$ & {\color{red}Uranker}$\uparrow$ & {\color{green}PAQ2PIQ}$\uparrow$ & {\color{blue}MUSIQ}$\uparrow$ & {\color{red}Uranker}$\uparrow$ & {\color{green}PAQ2PIQ}$\uparrow$ & {\color{blue}MUSIQ}$\uparrow$ & {\color{red}Uranker}$\uparrow$ \\
\hline
$d_2$=5.0 & 74.921 & \textbf{54.598} & \textbf{2.553} & \textbf{73.417} & 48.715 & 2.502 & \textbf{73.315} & 52.288 & \textbf{2.409} & 1.11\\
$d_2$=7.5 & \textbf{74.923} & 54.595 & 2.549 & 73.415 & \textbf{48.718} & \textbf{2.504} & 73.313 & \textbf{52.290} & 2.406 & 1.11\\

\end{tabular}}
\caption{Quantitative performance based on different $d_2$ settings. The Avg. Rank means the average metric performances compared with previous works in Tab.\ref{metric_comp}.}
\label{horizontal_depth_metric_comp}
\end{table*}

\subsection{Application: Boosting Downstream Tasks}

\begin{figure}[pos=!t]
      \centering
      \includegraphics[width=1.0\linewidth]{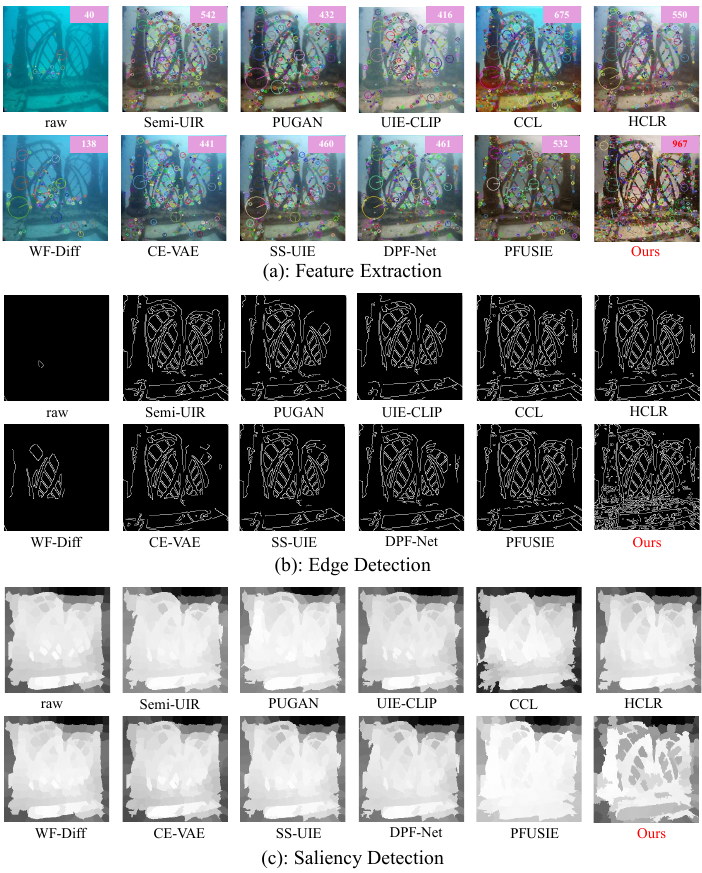}
      \caption{Examples of the downstream applications about feature extraction (a), edge (b) and saliency detection (c). The number in the images of subfigure (a) denotes the amount of detected features.}
      \label{ds_apps_1} 
    \end{figure}

\begin{figure}[pos=!t]
      \centering
      \includegraphics[width=1.0\linewidth]{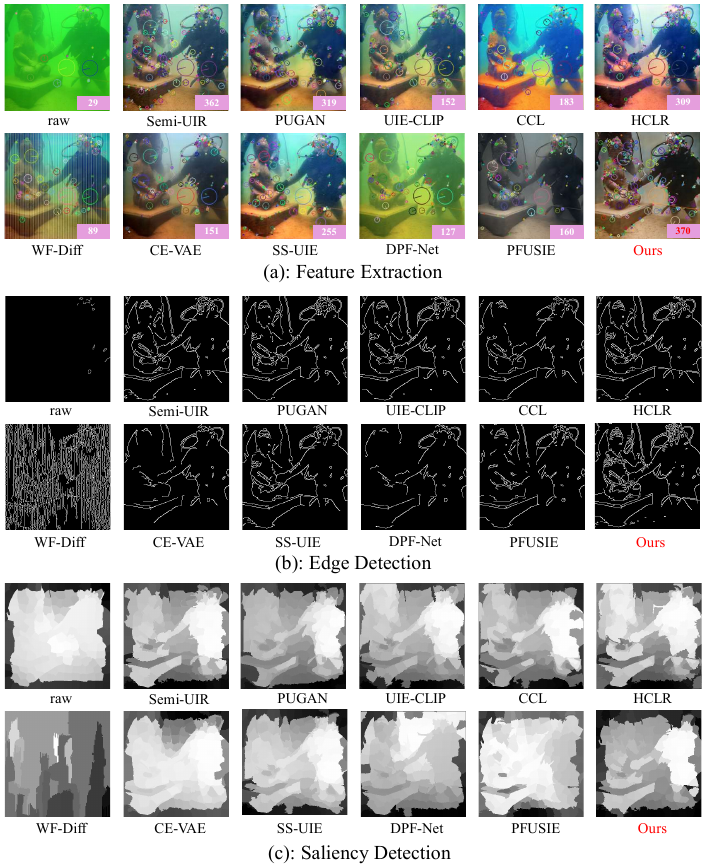}
      \caption{Examples of the downstream applications about feature extraction (a), edge (b) and saliency detection (c). The number in the images of subfigure (a) denotes the amount of detected features.}
      \label{ds_apps_2} 
    \end{figure}

\begin{figure}[pos=!t]
      \centering
      \includegraphics[width=1.0\linewidth]{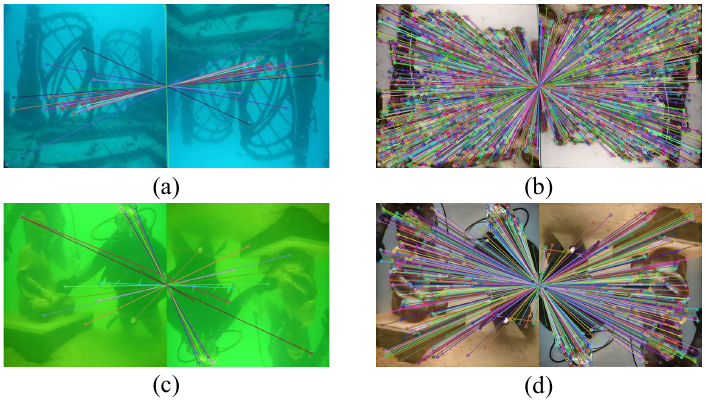}
      \caption{The result of feature matching by the raw underwater images (a), (c), and the corresponding images enhanced by the proposal (b), (d). }
      \label{ds_apps_4} 
    \end{figure}

A promising application is to use the enhanced images to improve the performance of downstream vision tasks such as feature detection, feature matching, edge extraction, and saliency detection. In this section, we provide severe examples of underwater images under challenging degradations to test the above four downstream tasks. Feature detection and matching are processed by SIFT \cite{lowe2004sift}, while the Canny edge detector \cite{canny2009canny} and the Manifold Ranking (MR) \cite{yang2013mrsaliency} are implemented for the detection of the edges and the saliency . As shown in Fig.\ref{ds_apps_1}, \ref{ds_apps_2}, where the subfigures (a), (b), and (c) are the comparison of the results between different methods in the detection of features, edge detection, and saliency detection, the underwater images enhanced by the proposed P2UIE achieve the best average performance for downstream tasks, including more detected features, richer extracted edges, and better alignment with the real salient region. This competitive result comes from the proposed transfer learning strategy: The noisy labels where the color bias and low contrast effect are not well solved are discarded. Instead, P2UIE resorts to multiple priors from relevant research fields for physics-based transfer learning to correctly improve image quality. In this way, the diffusion prior, real-world dense haze prior, and semantic prior are properly transferred to the model so that the generated results are with high contrast of the foreground part/smooth background part, rich and correct color pattern as desired in the downstream tasks. Fig.\ref{ds_apps_4} shows that the number of matching features improved significantly, where the second image is the flipped version of the raw underwater image at 180 degrees.

\subsection{Run-time Analysis and Limitations}

The average run-time comparison is given in the Tab.\ref{comp_ana}. Since the proposed method is based on a diffusion model with a large model size, its parameter count and FLOPs do not exhibit an advantage. However, the acceptable inference time (around 8.69Hz) and GPU memory consumption (3133 MB) already support the real-time implementation on the mobile underwater platforms equipped with computational graphics like Jetson Nano, and the leading performance of our proposal represents a reasonable trade-off for the computational resources, particularly in applications requiring high-quality images.

\begin{table}[pos=!t]
\centering
\scalebox{0.65}{
\begin{tabular}{lccccc}
\hline
Method & Param. (MB) & FLOPs (G) & Inf. Time (ms) & GPU Mem. (MB) & Avg. Rank \\
\hline
Semi-UIR & 1.68   & 39.91  & 1840 & 2135 & 3.33 \\
PUGAN    & 189.15 & 73.97  & 15   & 3599 & 6.00 \\
CCL      & 1.76   & 109.10 & 56   & 811  & 5,78\\
HCLR     & 19.55  & 398.36 & 190  & 1345 & 6.89\\
WF-Diff  & 100.55  & 1197.53    & 593  & 1165 & 8.11\\
CE-VAE   & 83.44  & 237.15 & 118  & 5537 & 8.00\\
SS-UIE   & 20.63  & 15.15  & 117  & 5005 & 7.44\\
DPF-Net  & 34.44  & 89.42  & 145  & 2483 & 7.78\\
PFUSIE   & 1622.36  &994.95 & 294.2 & 9981 & 4.44\\
UIE-CLIP & 3.15   & 10.46  & 33   & 2453 & 7.11\\
Ours     & 949.57  & 530.65  & 115  & 3133 & 1.11\\
\hline
\end{tabular}}  
\caption{The computational complexity per image among different methods. The Avg. Rank means the average performance ranking as given in Tab.\ref{metric_comp} (the smaller is better).}
\label{comp_ana}1
\end{table}

Although much progress has been made with the proposal, it still has drawbacks. For example, in the step of background noise suppression, the partial learning labels still require human intervention although the mask is designed to auto generate, as demonstrated in Fig.\ref{fuseproc}. This human-interactive step hinders the complete automation of the method, which is inconvenient for the collection of large-scale datasets. In addition, like other UIE methods, this approach is designed for single image enhancement, where long-term temporal consistency is not integrated into the training process as a constraint. This will affect its performance on downstream applications, such as visual odometry and SLAM. Future works will be dedicated to solving the above two issues. Despite that, the innovation of our proposal that leveraging the multi-priors from relevant vision tasks to cross-domain supervision as well as physics-based task decomposition is validated and shown to be superior to earlier methods.

\section{Conclusion}

This paper aims to design a P2-UIE that achieves SOTA performance without requiring the true labels that are hard to collect. Meanwhile, P2UIE avoids using the current pseudo-labels that are heavily noisy for supervised learning at the same time. To achieve this goal, multiple types of knowledge from relevant research fields are transferred for cross-domain supervision, including diffusion model prior, real-world hazy data prior, and semantic model prior. In addition, the UIE task is mainly decomposed into three steps, including global color correction, hazy removal, and background noise suppression. Each step focuses only on the local enhancement task to decrease the learning difficulty, and the physics-based decomposition further provides the theoretical soundness. The result analysis and the downstream applications support the effectiveness and significance of the proposed method, which is qualitatively and quantitatively superior to previous works. 

\section{Acknowledgment}

This work was jointly supported by the Research Grants Council of Hong Kong (25206524, 15212925), the National Natural Science Foundation of China (42301520), the Innovation and Technology Fund (PRP/068/23FX), the Seed Projects of Smart Cities Research Institute (P0051028, P0054511).

\bibliographystyle{cas-model2-names}

\bibliography{refs}

\end{document}